\documentclass[acmtog]{acmart}

\usepackage{graphicx}
\usepackage{amsmath}
\usepackage{booktabs}
\usepackage{multirow}
\usepackage{threeparttable}
\usepackage{siunitx}
\usepackage{booktabs} 
\usepackage{amsfonts}
\usepackage{graphicx}
\acmJournal{TOG}
\usepackage{multirow}
\usepackage{subcaption}
\usepackage{amsmath} 
\usepackage{graphicx}                  
\usepackage[utf8]{inputenc}            
\usepackage[T1]{fontenc}   
\usepackage{threeparttable}
\usepackage{siunitx}
\usepackage{lipsum} 
\usepackage{pdfpages}
\usepackage{tabularx}
\usepackage[ruled]{algorithm2e} 

\SetAlFnt{\small}
\SetAlCapFnt{\small}
\SetAlCapNameFnt{\small}

\SetAlCapHSkip{0pt}
\usepackage{enumitem}

\usepackage{subcaption}
\acmBooktitle{}

\copyrightyear{2025}
\acmYear{2025}
\setcopyright{cc}
\setcctype{by}
\setcctype{by}
\acmConference[SA Conference Papers '25]{SIGGRAPH Asia 2025 Conference
Papers}{December 15--18, 2025}{Hong Kong, Hong Kong}
\acmBooktitle{SIGGRAPH Asia 2025 Conference Papers (SA Conference Papers
'25), December 15--18, 2025, Hong Kong, Hong
Kong}
\begin{document}

\newcommand{\targetimg}{\mathbf{T}}  
\newcommand{\tindex}{j} 
\newcommand{\pt}{x} 
\newcommand{\viewdir}{v} 
\newcommand{\handpose}{p}  
\newcommand{\signeddist}{d}  
\newcommand{\featureop}{f_{32}} 

\newcommand{\etal}{\textit{et al.}}

\def\reta{{\textnormal{$\eta$}}}
\def\ra{{\textnormal{a}}}
\def\rb{{\textnormal{b}}}
\def\rc{{\textnormal{c}}}
\def\rd{{\textnormal{d}}}
\def\re{{\textnormal{e}}}
\def\rf{{\textnormal{f}}}
\def\rg{{\textnormal{g}}}
\def\rh{{\textnormal{h}}}
\def\ri{{\textnormal{i}}}
\def\rj{{\textnormal{j}}}
\def\rk{{\textnormal{k}}}
\def\rl{{\textnormal{l}}}
\def\rn{{\textnormal{n}}}
\def\ro{{\textnormal{o}}}
\def\rp{{\textnormal{p}}}
\def\rq{{\textnormal{q}}}
\def\rr{{\textnormal{r}}}
\def\rs{{\textnormal{s}}}
\def\rt{{\textnormal{t}}}
\def\ru{{\textnormal{u}}}
\def\rv{{\textnormal{v}}}
\def\rw{{\textnormal{w}}}
\def\rx{{\textnormal{x}}}
\def\ry{{\textnormal{y}}}
\def\rz{{\textnormal{z}}}

\def\rvepsilon{{\mathbf{\epsilon}}}
\def\rvtheta{{\mathbf{\theta}}}
\def\rva{{\mathbf{a}}}
\def\rvb{{\mathbf{b}}}
\def\rvc{{\mathbf{c}}}
\def\rvd{{\mathbf{d}}}
\def\rve{{\mathbf{e}}}
\def\rvf{{\mathbf{f}}}
\def\rvg{{\mathbf{g}}}
\def\rvh{{\mathbf{h}}}
\def\rvu{{\mathbf{i}}}
\def\rvj{{\mathbf{j}}}
\def\rvk{{\mathbf{k}}}
\def\rvl{{\mathbf{l}}}
\def\rvm{{\mathbf{m}}}
\def\rvn{{\mathbf{n}}}
\def\rvo{{\mathbf{o}}}
\def\rvp{{\mathbf{p}}}
\def\rvq{{\mathbf{q}}}
\def\rvr{{\mathbf{r}}}
\def\rvs{{\mathbf{s}}}
\def\rvt{{\mathbf{t}}}
\def\rvu{{\mathbf{u}}}
\def\rvv{{\mathbf{v}}}
\def\rvw{{\mathbf{w}}}
\def\rvx{{\mathbf{x}}}
\def\rvy{{\mathbf{y}}}
\def\rvz{{\mathbf{z}}}

\def\erva{{\textnormal{a}}}
\def\ervb{{\textnormal{b}}}
\def\ervc{{\textnormal{c}}}
\def\ervd{{\textnormal{d}}}
\def\erve{{\textnormal{e}}}
\def\ervf{{\textnormal{f}}}
\def\ervg{{\textnormal{g}}}
\def\ervh{{\textnormal{h}}}
\def\ervi{{\textnormal{i}}}
\def\ervj{{\textnormal{j}}}
\def\ervk{{\textnormal{k}}}
\def\ervl{{\textnormal{l}}}
\def\ervm{{\textnormal{m}}}
\def\ervn{{\textnormal{n}}}
\def\ervo{{\textnormal{o}}}
\def\ervp{{\textnormal{p}}}
\def\ervq{{\textnormal{q}}}
\def\ervr{{\textnormal{r}}}
\def\ervs{{\textnormal{s}}}
\def\ervt{{\textnormal{t}}}
\def\ervu{{\textnormal{u}}}
\def\ervv{{\textnormal{v}}}
\def\ervw{{\textnormal{w}}}
\def\ervx{{\textnormal{x}}}
\def\ervy{{\textnormal{y}}}
\def\ervz{{\textnormal{z}}}

\def\rmA{{\mathbf{A}}}
\def\rmB{{\mathbf{B}}}
\def\rmC{{\mathbf{C}}}
\def\rmD{{\mathbf{D}}}
\def\rmE{{\mathbf{E}}}
\def\rmF{{\mathbf{F}}}
\def\rmG{{\mathbf{G}}}
\def\rmH{{\mathbf{H}}}
\def\rmI{{\mathbf{I}}}
\def\rmJ{{\mathbf{J}}}
\def\rmK{{\mathbf{K}}}
\def\rmL{{\mathbf{L}}}
\def\rmM{{\mathbf{M}}}
\def\rmN{{\mathbf{N}}}
\def\rmO{{\mathbf{O}}}
\def\rmP{{\mathbf{P}}}
\def\rmQ{{\mathbf{Q}}}
\def\rmR{{\mathbf{R}}}
\def\rmS{{\mathbf{S}}}
\def\rmT{{\mathbf{T}}}
\def\rmU{{\mathbf{U}}}
\def\rmV{{\mathbf{V}}}
\def\rmW{{\mathbf{W}}}
\def\rmX{{\mathbf{X}}}
\def\rmY{{\mathbf{Y}}}
\def\rmZ{{\mathbf{Z}}}

\def\ermA{{\textnormal{A}}}
\def\ermB{{\textnormal{B}}}
\def\ermC{{\textnormal{C}}}
\def\ermD{{\textnormal{D}}}
\def\ermE{{\textnormal{E}}}
\def\ermF{{\textnormal{F}}}
\def\ermG{{\textnormal{G}}}
\def\ermH{{\textnormal{H}}}
\def\ermI{{\textnormal{I}}}
\def\ermJ{{\textnormal{J}}}
\def\ermK{{\textnormal{K}}}
\def\ermL{{\textnormal{L}}}
\def\ermM{{\textnormal{M}}}
\def\ermN{{\textnormal{N}}}
\def\ermO{{\textnormal{O}}}
\def\ermP{{\textnormal{P}}}
\def\ermQ{{\textnormal{Q}}}
\def\ermR{{\textnormal{R}}}
\def\ermS{{\textnormal{S}}}
\def\ermT{{\textnormal{T}}}
\def\ermU{{\textnormal{U}}}
\def\ermV{{\textnormal{V}}}
\def\ermW{{\textnormal{W}}}
\def\ermX{{\textnormal{X}}}
\def\ermY{{\textnormal{Y}}}
\def\ermZ{{\textnormal{Z}}}

\def\vzero{{\bm{0}}}
\def\vone{{\bm{1}}}
\def\vmu{{\bm{\mu}}}
\def\vtheta{{\bm{\theta}}}
\def\va{{\bm{a}}}
\def\vb{{\bm{b}}}
\def\vc{{\bm{c}}}
\def\vd{{\bm{d}}}
\def\ve{{\bm{e}}}
\def\vf{{\bm{f}}}
\def\vg{{\bm{g}}}
\def\vh{{\bm{h}}}
\def\vi{{\bm{i}}}
\def\vj{{\bm{j}}}
\def\vk{{\bm{k}}}
\def\vl{{\bm{l}}}
\def\vm{{\bm{m}}}
\def\vn{{\bm{n}}}
\def\vo{{\bm{o}}}
\def\vp{{\bm{p}}}
\def\vq{{\bm{q}}}
\def\vr{{\bm{r}}}
\def\vs{{\bm{s}}}
\def\vt{{\bm{t}}}
\def\vu{{\bm{u}}}
\def\vv{{\bm{v}}}
\def\vw{{\bm{w}}}
\def\vx{{\bm{x}}}
\def\vy{{\bm{y}}}
\def\vz{{\bm{z}}}

\title{Audio-Driven Universal Gaussian Head Avatars}

\author{Kartik Teotia}
\email{kteotia@mpi-inf.mpg.de}
\orcid{0009-0007-6985-7159}
\affiliation{%
  \institution{Max Planck Institute for Informatics and Saarland Informatics Campus}
  \country{Germany}
}

\author{Helge Rhodin}
\email{hrhodin@mpi-inf.mpg.de}
\orcid{0000-0003-2692-0801}
\affiliation{%
  \institution{Max Planck Institute for Informatics}
  \country{Germany}
}

\author{Mohit Mendiratta}
\email{mmendiratta@mpi-inf.mpg.de}
\orcid{0009-0001-5577-157X}
\affiliation{%
  \institution{Max Planck Institute for Informatics and Saarland Informatics Campus}
  \country{Germany}
}
\author{Hyeongwoo Kim}
\email{hyeongwoo.kim@imperial.ac.uk}
\orcid{0000-0002-9685-2579}
\affiliation{%
  \institution{Imperial College London}
  \country{United Kingdom}
}

\author{Marc Habermann}
\email{mhaberma@mpi-inf.mpg.de}
\orcid{0000-0003-3899-7515}
\affiliation{%
  \institution{Max Planck Institute for Informatics and Saarland Informatics Campus}
  \country{Germany}
}

\author{Christian Theobalt}
\email{theobalt@mpi-inf.mpg.de}
\orcid{0000-0001-6104-6625}
\affiliation{%
  \institution{Max Planck Institute for Informatics and Saarland Informatics Campus}
  \country{Germany}
}

\begin{abstract}
   %
%
We introduce the first method for audio-driven universal photorealistic avatar synthesis, combining a person-agnostic speech model with our novel Universal Head Avatar Prior (UHAP). 
UHAP is trained on cross-identity multi-view videos. 
In particular, our UHAP is supervised with neutral scan data, enabling it to capture the identity-specific details at high fidelity. 
In contrast to previous approaches, which predominantly map audio features to geometric deformations only while ignoring audio-dependent appearance variations, our universal speech model directly maps raw audio inputs into the UHAP latent expression space.
This expression space inherently encodes, both, geometric and appearance variations. 
For efficient personalization to new subjects, we employ a monocular encoder, which enables lightweight regression of dynamic expression variations across video frames. 
By accounting for these expression-dependent changes, it enables the subsequent model fine-tuning stage to focus exclusively on capturing the subject's global appearance and geometry. 
Decoding these audio-driven expression codes via UHAP generates highly realistic avatars with precise lip synchronization and nuanced expressive details, such as eyebrow movement, gaze shifts, and realistic mouth interior appearance as well as motion. 
Extensive evaluations demonstrate that our method is not only the first generalizable audio-driven avatar model that can account for detailed appearance modeling and rendering, but it also outperforms competing (geometry-only) methods across metrics measuring lip-sync accuracy, quantitative image quality, and perceptual realism. Webpage: \href{https://kartik-teotia.github.io/UniGAHA/}{\textcolor{blue}{https://kartik-teotia.github.io/UniGAHA/}}
%
%
\end{abstract}

\keywords{Audio-driven animation, Gaussian Head Avatars}  
\begin{teaserfigure}
\centering
  \includegraphics[width=\textwidth]{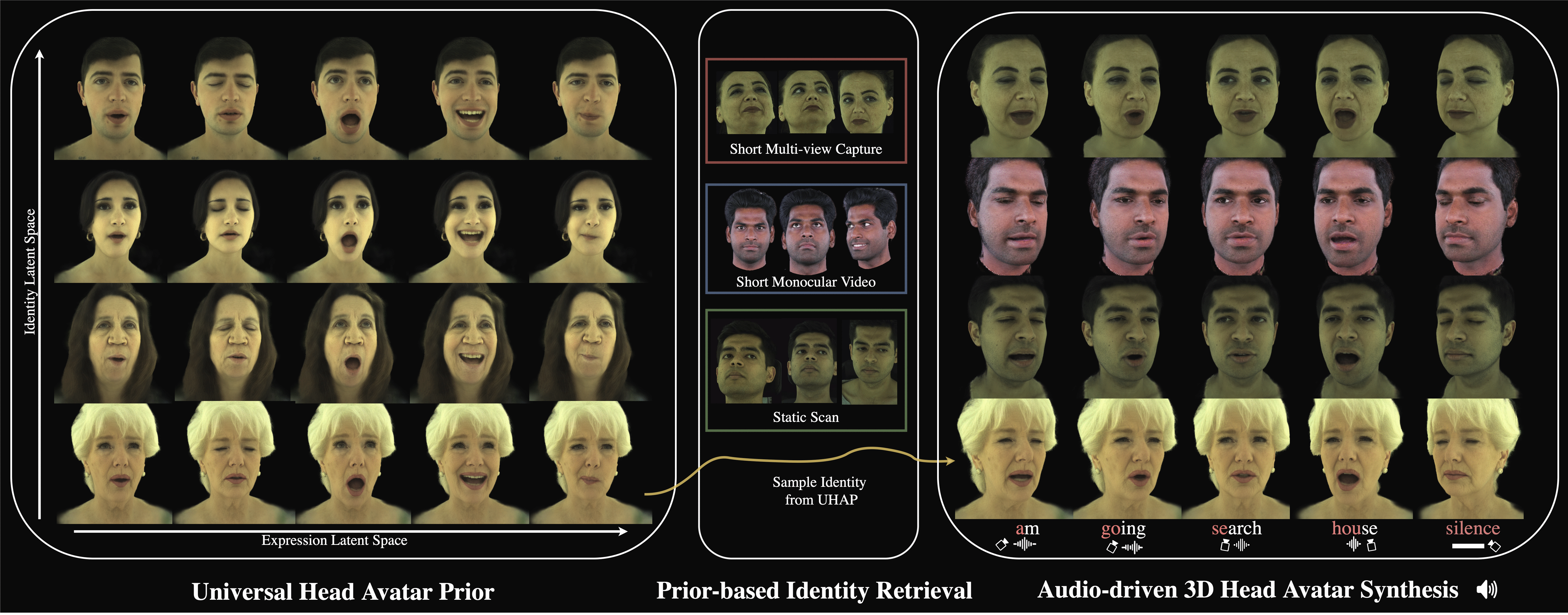}
  \caption{We present a new method for audio-driven photorealistic 3D head avatar synthesis with a Universal Head Avatar Prior (\textbf{UHAP}). (Left) We learn a Universal Head Avatar Prior from a diverse dataset, capturing rich facial geometry and appearance across multiple identities (rows) and dynamic expressions (columns). (Center) Given minimal data for a new subject—whether a short multi‐view capture, a monocular video clip, or a single static scan—we retrieve a personalized identity from the \textbf{UHAP}. (Right) Conditioning the retrieved identity on an arbitrary speech waveform yields high‐fidelity, lip-synced full-face animations that faithfully preserve identity and expression dynamics across multiple subjects.}
  \label{fig:teaser}
\end{teaserfigure}

\maketitle

%
%
\section{Introduction} \label{sec:intro}
Synthesizing photorealistic 3D head avatars, which can be driven solely from speech, presents a compelling avenue for applications ranging from virtual communication to digital entertainment \cite{dummy_audio_avatar_importance1}. 
The goal is to generate accurate lip synchronization along with expressive facial motion and, crucially, realistic visual appearance, while also ensuring temporal and view-point consistency. 
Achieving this using only speech as input is particularly valuable due to the lightweight sensor modality required to capture these audio signals. 
%
%
\par
A primary challenge lies in generating photorealistic appearance synchronized with accurate 3D facial motion derived from speech, while also ensuring the model generalizes to novel identities either by sampling a novel identity or by finetuning on few shot data of a real person.
Recent advancements in audio-driven video synthesis, such as VASA-1\cite{xu2024vasa1lifelikeaudiodriventalking}, have demonstrated impressive results in generating lifelike talking faces. These methods leverage the power of diffusion-based models for mapping audio features to a video latent space to create highly realistic animations. However, these state-of-the-art approaches primarily operate in 2D, synthesizing video frames that, while visually compelling, lack the underlying 3D structure necessary for applications requiring free-viewpoint rendering.
Achieving combined realism in motion and appearance in 3D remains difficult, especially without prohibitive per-person requirements like extensive multi-view capture sessions or hours of subject-specific training time \cite{aneja2024gaussianspeech,Richard_2021_WACV}.
%
%
\par 
Traditional approaches to audio-driven 3D animation utilize geometric representations like 3D Morphable Models (3DMMs) \cite{dummy_audio2face_3dmm_1,dummy_faceformer_2022} or artist designed template meshes \cite{audiodrivenfacialanimationbyjointendtoen}. 
While suitable for controlling basic 3D shape and motion, they face a key limitation: 
they do not model dynamic textures and view-dependent appearance directly from the audio signal. 
This deficiency makes it particularly difficult to realistically render regions such as the mouth interior or gaze shifts during speech.
Consequently, the visual results often fall short of the photorealism required by many modern applications 
%
%
\par 
Techniques employing modern photorealistic representations like Neural Radiance Fields (NeRFs) \cite{dummy_nerf_ref} or 3D Gaussian Splatting (3DGS) \cite{dummy_3dgs} excel at capturing appearance for static scenes or controlled dynamic captures. 
However, applying them directly to audio-driven animation across diverse identities often involves costly per-subject optimization or training \cite{aneja2024gaussianspeech, Richard_2021_WACV, ng2024audiophotorealembodimentsynthesizing}, requiring significant computation time (hours to days) and large amounts of per-person data, thus, hindering the creation of universal and readily deployable models. 
Moreover, many recent audio-driven methods, even when using the powerful diffusion models \cite{dummy_diffposetalk_ref, dummy_media2face_ref}, still primarily focus on driving intermediate geometry, thereby inheriting the appearance and expressiveness limitations of those representations.
%
%
\par 
Our work addresses these limitations through a novel framework centered around three key technical contributions.
First, we construct a Universal Head Avatar Prior (UHAP) based on 3D Gaussian Splatting (3DGS) \cite{3DGS}. 
This prior is trained on large-scale multi-view dynamic videos from studio captures, and critically incorporates supervision from neutral scan data to preserve identity-specific details during training. 
The resulting UHAP learns an avatar representation representation with effectively disentangled latent spaces for identity and expression. 
Second, unlike prior work mapping audio to intermediate geometry, we leverage a diffusion-based speech model that maps raw audio features directly into the UHAP's expression space. 
A key aspect of our approach is that these predicted latent parameters explicitly encode both geometry (e.g. mouth motion) and appearance (e.g. gaze shifts) variations.
Third, we enable efficient personalization of the UHAP to new subjects from sparse data, enabling practical applications such as driving a subject from a single static capture, or a short monocular video. 
Key to our adaptation process is a generalized monocular image encoder that estimates and factors out expression dynamics within the video frames.
Thereby our monocular finetuning stage captures the target identity's global appearance and geometry.
Importantly, our monocular image encoder does not rely on acquiring explicit geometry and appearance tracking. 
Decoding the audio-driven expression codes via the personalized UHAP yields the final photorealistic avatars with high-fidelity facial motion and naturally synchronized appearance changes.
%
%
\par 
In summary, our key contributions are:
\begin{itemize}
    \item A universal framework for audio-driven and photorealistic 3D Gaussian head avatar generation allowing unconditional identity sampling as well as few-shot identity finetuning while preserving an audio-driven expression latent space.
    \item To this end, we first introduce a Universal Head Avatar Prior (UHAP), which effectively disentangles identity and expression latent spaces while ensuring high-fidelity synthesis thanks to our neutral texture formulation.
    \item A diffusion-based speech model that maps input audio to the UHAP's expression latent space, which enables driving of the underlying 3D facial geometry and appearance. This mapping to a 3D avatar prior ensures view- and-identity consistent facial animations.
    \item Our monocular expression encoder facilitates a variety of few-shot identity finetuning applications such as finetuning the identity solely on a static scan or a short monocular video.
\end{itemize}
To the best of our knowledge, our work is the first that demonstrates generalization across individuals while also enabling audio-driven appearance synthesis.
Our evaluation further demonstrates that we outperform geometry-only baselines in terms of audio-visual synchronization as well as visual appearance.
%
%

%
%
\section{Related Work} \label{sec:related_work}
For universal avatar models to be practical for widespread adoption, they must satisfy three key criteria: 
they should accurately represent diverse identities, capture nuanced speech-driven expressions, and enable easy personalization from sparse observations. 
In what follows, we discuss prior works according to these criteria, highlighting their strengths and identifying key limitations. 
%
%
\subsection{Speech-driven Geometric Facial Representations}
The generation of 3D facial animation from audio has a rich history, with 3D Morphable Models (3DMMs)~\cite{dummy_blanz_vetter_1999, li2017_flame} offering a generalized parametric framework for representing facial geometry and appearance. 
Previous approaches often involved mapping acoustic features to the parameters of these 3DMMs to achieve speech-driven animation~\cite{diffposetalkspeechdrivenstylistic3dfacia,voice2faceaudiodrivenfacialandtongueriga,emotalkspeechdrivenemotionaldisentanglem,Danecek_2022_CVPR}. 
However, these models are often constrained by the expressive capacity inherent in the 3DMM's low-dimensional Principal Component Analysis (PCA) parameters, which can struggle to capture the full range of subtle, high-fidelity dynamics. 
Recognizing these limitations, other approaches have focused on directly modeling more detailed geometric deformations~\cite{dummy_meshtalk_2021, dummy_faceformer_2022}. 
More recently, deep generative approaches, particularly diffusion models, have gained traction in this domain~\cite{facediffuserspeechdriven3dfacialanimatio, diffposetalkspeechdrivenstylistic3dfacia, media2facecospeechfacialanimationgenerat}. 
While powerful, many of these diffusion-based methods still focus on predicting parameters for established representations like 3DMMs or geometric latent models~\cite{aneja2023facetalk}. 
However, a key limitation across many speech-driven geometric representations is their inability to directly model or synthesize nuanced, speech-correlated appearance changes, such as subtle gaze shifts, or deforming mouth interior.
Addressing this gap, our approach synthesizes expression latents of our Universal Head Avatar Prior (UHAP), which jointly encodes, both, the subject-agnostic geometry-dependent expression changes and dynamic appearance.
%
%
\subsection{Speech-driven Appearance Methods}
Integrating realistic, dynamic appearance with speech-driven animation is crucial for photorealism but remains challenging. 
Early efforts primarily focused on 2D audio-driven facial animation from monocular RGB videos~\cite{dummy_chen_2018_lipmovements,dummy_guan_2023_stylesync}. 
These 2D methods, while achieving plausible lip sync, operate in pixel space, and, thus, they can neither achieve 3D consistency nor they support free-viewpoint rendering. 
Transitioning to 3D, many recent efforts leveraging Neural Radiance Fields (NeRF) for talking head synthesis from monocular video have shown impressive photorealism, such as AD-NeRF~\cite{dummy_adnerf_2021_guo} and GeneFace~\cite{dummy_geneface_2023_ye}, but these are often person-specific and require per-subject optimization.
Other works like RAD-NeRF~\cite{dummy_radnerf_2022_tang} and ER-NeRF~\cite{dummy_ernerf_2023_li} focus on efficient, real-time synthesis from audio for personalized avatars. 
Audio-driven codec avatars, as explored in~\cite{Richard_2021_WACV, ng2024audiophotorealembodimentsynthesizing}, can produce high-fidelity personalized results but also operate on a per-subject basis. 
Similarly, GaussianSpeech~\cite{aneja2024gaussianspeech} achieves detailed, personalized audio-driven avatars using 3D Gaussian Splatting by learning expression-dependent color and dynamic wrinkles, but is tailored to individual subjects.
TexTalker~\cite{dummy_textalker_2025_li}, a concurrent work to ours, generates dynamic textures aligned with speech-driven facial motion, using a high-resolution 4D dataset. 
It proposes a diffusion-based framework to simultaneously generate facial motions and dynamic textures from speech for personalized avatars. 
While TexTalker addresses dynamic textures, our work distinguishes itself by aiming for a universal prior that holistically controls, both, geometry and the broader appearance attributes captured by 3D Gaussians, not limited to the tracked 2D texture maps, and allows for efficient adaptation to new individuals.

\subsection{Gaussian Avatar Representations}

Recent works leveraging 3D Gaussian Splatting \cite{3DGS} have introduced several powerful representations for creating personalized, animatable head avatars. Foundational approaches such as GaussianAvatars \cite{qian2024gaussianavatars} rig 3D Gaussians directly to the FLAME model \cite{li2017_flame}, while others learn to deform a canonical set of Gaussians conditioned on global expression parameters \cite{teotiagaussian,saito2024rgca,giebenhain2024npga}. ScaffoldAvatar \cite{aneja2025scaffoldavatar} achieves high fidelity rendering of faces using localized patch-based expressions. Gaussian Blendshapes \cite{ma2024gaussianblendshapes} introduce an explicit blendshape formulation, where a full set of expression bases directly modulates Gaussian parameters for facial animation. RGBAvatar \cite{li2025rgbavatar} streamlines this design by predicting a reduced blendshape basis from FLAME expressions, yielding a more compact representation that supports efficient online training and real-time rendering. Specialized models like GaussianSpeech \cite{aneja2024gaussianspeech} animates subject-specific avatars by using a transformer model to predict audio-driven mesh deformations, which then drive the final 3D Gaussian representation. While these works provide powerful representations for creating high-fidelity personalized Gaussian avatars, often requiring extensive per-subject data and training, our approach introduces a Universal Head Avatar Prior (UHAP). This generalizable, person-agnostic model learns a disentangled, cross-identity latent space for facial expressions that enables high-fidelity animation from multiple modalities, such as an audio stream or a driving video, and supports efficient, few-shot personalization from limited input data for a new subject like a short monocular video or a static capture from multiple views.
\subsection{Universal Avatar Priors}

Universal avatar models, capable of representing diverse identities and expressions within a unified framework, are pivotal for enabling generalization to new individuals.
Significant progress has been made in this domain. For instance, some recent universal priors focus on achieving relighting capabilities alongside expressive control; URAvatar~\cite{li2024uravatar} and VRMM~\cite{yang2024vrmm} are notable examples that allow for avatars to be rendered under novel illumination conditions, with VRMM also emphasizing volumetric representations built from data captured under controlled lighting. Other efforts, such as Authentic Volumetric Avatars by Cao et al.~\cite{cao2022_authentic}, have pushed the boundaries of creating high-fidelity, animatable volumetric avatars from inputs like phone scans.  While these works show promising results in creating generalizable and high-fidelity avatars, adapting them to new, unseen identities can present challenges. For example, approaches such as those by Cao et al.~\cite{cao2022_authentic} and Li et al.~\cite{li2024uravatar} (URAvatar) may involve extensive fine-tuning or the acquisition of dynamically tracked non-rigid facial geometry~\cite{grassal2022_neural-head}. Additionally, many recent approaches based on 3D Gaussian Splatting map inputs to lower-dimensional expression spaces derived from, or aligned with, parametric models~\cite{zheng2024headgap, xu2024gphm}, which can constrain the overall expressiveness. Avat3r \cite{kirschstein2025avat3rlargeanimatablegaussian} presents a feed-forward method for avatar creation using phone scans; however, its animation is driven by signals restricted to studio-tracked captures. Our Universal Head Avatar Prior (UHAP), embedded within a 3D Gaussian Splatting framework, presents a framework for lightweight personalization as well animation. It achieves lightweight personalization—facilitated by our image encoder that effectively disentangles dynamic subject-specific variations from global appearance and expression nuances—and, critically, learns a richer latent expression space directly from high-quality studio data, moving beyond the constraints of predefined parametric models. This learned expression space directly modulates 3D Gaussian properties for animation from lightweight input signals such as audio or monocular images.
%
%
%
\section{Method} \label{sec:method}
%
\begin{figure*}[htb]
\centering
\includegraphics[width=\textwidth]{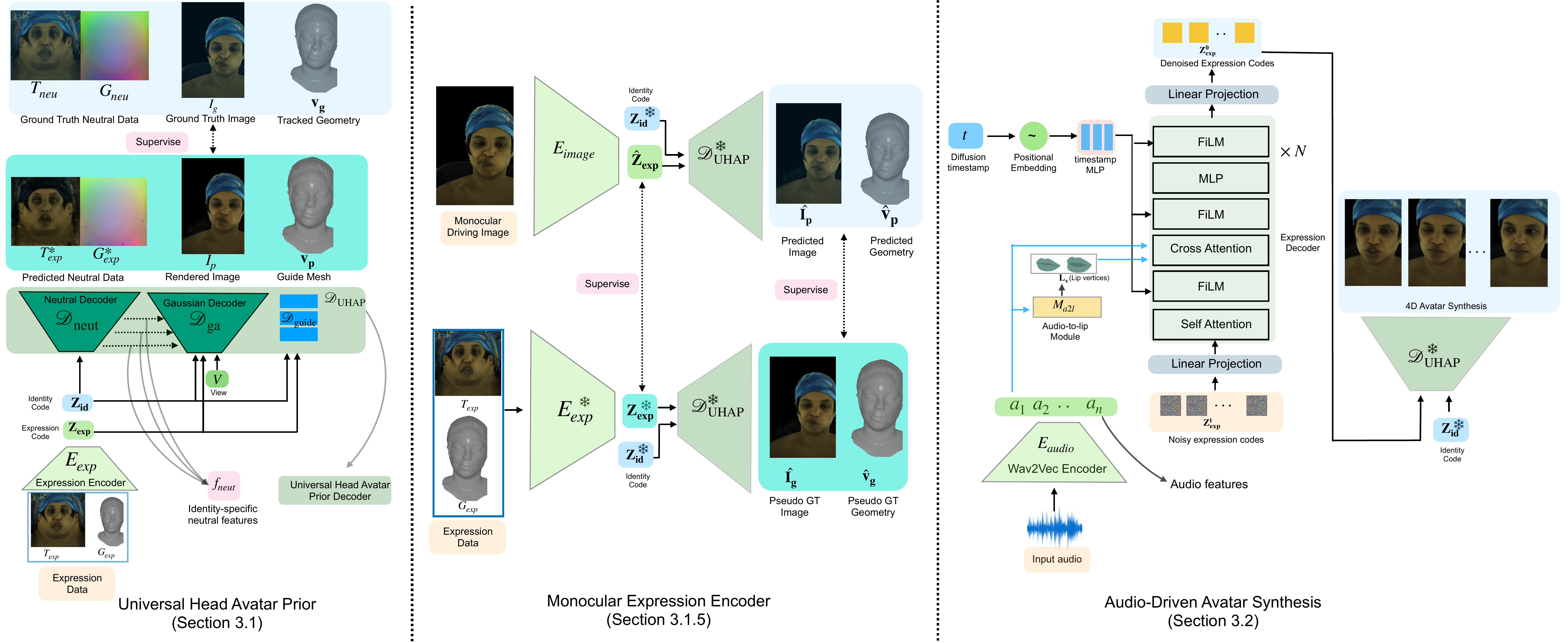}
\caption{
Overview of our Audio-Driven Universal Gaussian Avatar pipeline. 
This figure illustrates the three main stages: 
(\textbf{Sec.~\ref{sec:uhap}}) Universal Head Avatar Prior (UHAP) Training: A universal decoder $\mathcal{D}_\mathrm{UHAP}$ is trained on multi-identity, multi-view data to learn disentangled latent codes for identity ($\mathbf{Z}_\mathrm{id}$) and expression ($\mathbf{Z}_\mathrm{exp}$).
(\textbf{Sec.~\ref{sec:mono_encoder}}) Monocular Expression Encoder Training: An image encoder $E_\mathrm{image}$ predicts expression codes $\mathbf{\hat{Z}}_\mathrm{exp}$ from single images, supervised by $E_\mathrm{exp}$ (from UHAP) and reconstruction losses using pseudo-ground truth data ($\hat{I}_\mathrm{g}$, $\mathbf{\hat{v}}_\mathrm{g}$).
(\textbf{Sec.~\ref{sec:audio_synthesis}}) Audio-Driven Avatar Synthesis: A diffusion model generates expression code sequences $\mathbf{Z}^{0}_\mathrm{exp}$ from audio features which, combined with $\mathbf{Z}_\mathrm{id}$, drive the frozen $\mathcal{D}_\mathrm{UHAP}$ to synthesize the final animation.
}
\label{fig:overview}
\end{figure*}
%
Our method synthesizes photorealistic, audio-driven 3D talking head avatars, designed for cross-identity generalization and efficient personalization. It integrates a new high-fidelity Universal Head Avatar Prior (UHAP), built upon 3D Gaussian Splatting (3DGS)~\cite{3DGS}, with a universal diffusion-based speech model. This approach facilitates a direct mapping from audio features to a latent space encoding, both, expression-dependent geometry and appearance variations. Addressing the common challenge of acquiring large-scale, perfectly aligned multi-modal data (including synchronized audio, dynamic 3D geometry, and appearance across diverse identities), our framework is designed to effectively utilize varied data sources; for example, the universal prior can be trained on multi-view data even if it lacks corresponding audio, while the audio-driven aspects are learned subsequently with a diffusion model in the learned latent space.
Notably, personalization is possible from sparse subject-specific video or a static scan by using a monocular expression encoding and an optimization of an identity code and decoder fine-tuning, as outlined in Fig.~\ref{fig:overview}. The primary stages are: UHAP training (Sec.~\ref{sec:uhap}), learning a monocular expression encoder (Sec.~\ref{sec:mono_encoder}), and audio-driven avatar synthesis (Sec.~\ref{sec:audio_synthesis}), followed by a personalization stage for new subjects (Sec.~\ref{sec:personalization}). 
At inference, our method requires input audio (processed into features) to drive the personalized UHAP decoder. 
%
%
\subsection{Universal Head Avatar Prior (UHAP)} \label{sec:uhap}
The UHAP is our core model for synthesizing 3D Gaussian head avatars.
Synthesizing realistic, drivable 3D head avatars, especially from sparse inputs or solely audio, is an inherently ill-posed problem; UHAP addresses this by serving as a powerful, learned prior that constrains the synthesis process to enable high-fidelity and generalizable results.
It is conditioned on the identity code $\mathbf{Z}_\mathrm{id} \in \mathbb{R}^{D_\mathrm{id}}$ and an expression code $\mathbf{Z_\mathrm{exp}} \in \mathbb{R}^{D_\mathrm{exp}}$.
The identity code $\mathbf{Z}_\mathrm{id}$ aims to capture subject-specific canonical geometry and appearance, while $\mathbf{Z}_\mathrm{exp}$ controls facial deformations and associated appearance changes.
The UHAP is trained on dense multi-view video data from multiple subjects in the Ava-256 dataset~\cite{martinez2024codec}, which includes registered neutral 3D scans for each subject.
Unlike frameworks that encode pre-acquired neutral assets for new identities~\cite{li2024uravatar,cao2022_authentic}, our UHAP incorporates a Neutral Decoder component (Sec.~\ref{sec:uhap_decoder}).
This network learns to inject identity-specific features---derived from the neutral scan data during UHAP training and conditioned on $\mathbf{Z}_\mathrm{id}$---into the main avatar decoder.
This architecture promotes high-fidelity rendering.
Furthermore, for new, unseen identities, it allows for efficient fine-tuning using sparse data, such as a single static scan (Sec.~\ref{sec:personalization}).
Critically, our streamlined personalization strategy sidesteps an expensive and time-consuming precomputation; it does not necessitate non-rigid registration of the input dynamic or static data for these new subjects as is the case with \cite{li2024uravatar,cao2022_authentic}.
%
\subsubsection{Representation} \label{sec:uhap_repr}
We represent each avatar as a collection of $N_g = 256k$ 3D Gaussian primitives $\{ g_k \}_{k=1}^{N_g}$.
Each Gaussian $g_k = \{ \mathbf{t}_k \in \mathbb{R}^3, \mathbf{q}_k \in \mathbb{R}^4, \mathbf{s}_k \in \mathbb{R}^3_+, o_k \in \mathbb{R}_+, \mathbf{c}_k \in \mathbb{R}^{D_c} \}$ is defined by its center position $\mathbf{t}_k$, rotation as a unit quaternion $\mathbf{q}_k$, anisotropic scale $\mathbf{s}_k$, opacity $o_k$, and $D_c$ spherical harmonics (SH) coefficients encoding the color $\mathbf{c}_k$.
The rotation $\mathbf{q}_k$ and scale $\mathbf{s}_k$ together define the 3D Gaussian's covariance matrix.
Images $I$ are rendered differentiably from these primitives using the Gaussian rasterizer $\mathcal{R}(\{ g_k \}_{k=1}^{N_g})$, as proposed by \citet{3DGS}.
%
%
\subsubsection{Expression Encoder} \label{sec:uhap_exp_enc}
A variational autoencoder (VAE)~\cite{Kingma2013Auto}, $E_\mathrm{exp}$, learns the expression manifold.
The inputs to this encoder are UV-parameterized texture data ($T$) and geometry data ($G$).
To focus on expression-specific changes, $E_\mathrm{exp}$ processes the differences:
$\Delta T_\mathrm{exp} = T_\mathrm{exp} - T_\mathrm{neu}$ and
$\Delta G_\mathrm{exp} = G_\mathrm{exp} - G_\mathrm{neu}$.
These represent the deviations of the dynamic expression state ($T_\mathrm{exp}, G_\mathrm{exp}$) from a corresponding neutral state ($T_\mathrm{neu}, G_\mathrm{neu}$).
The encoder maps these differences into the parameters (mean $\boldsymbol{\mu}_\mathrm{exp}$ and standard deviation $\boldsymbol{\sigma}_\mathrm{exp}$) of a multivariate Gaussian distribution:
%
\begin{equation} \label{eq:exp_encoder}
\boldsymbol{\mu}_\mathrm{exp}, \boldsymbol{\sigma}_\mathrm{exp} = E_\mathrm{exp}(\Delta T_\mathrm{exp}, \Delta G_\mathrm{exp}; \Phi_{E_\mathrm{exp}})
\end{equation}
%
The expression code $\mathbf{Z}_\mathrm{exp} \in \mathbb{R}^{D_\mathrm{exp}}$ ($D_\mathrm{exp}=256$) is then sampled using the reparameterization trick~\cite{Kingma2013Auto}: $\mathbf{Z}_\mathrm{exp} = \boldsymbol{\mu}_\mathrm{exp} + \boldsymbol{\sigma}_\mathrm{exp} \cdot \boldsymbol{\epsilon}$, where $\boldsymbol{\epsilon} \sim \mathcal{N}(0, I)$.
%
\subsubsection{UHAP Decoder} \label{sec:uhap_decoder}
The UHAP Decoder, $\mathcal{D}_\mathrm{UHAP}$, synthesizes the full 3D Gaussian avatar conditioned on the identity code $\mathbf{Z}_\mathrm{id}$ and expression code $\mathbf{Z}_\mathrm{exp}$. It comprises three main components, each with its own set of learnable parameters denoted by $\Phi_{(\cdot)}$.
(1) A Neutral Decoder $\mathcal{D}_\mathrm{neut}$ processes $\mathbf{Z}_\mathrm{id}$ to produce identity-specific feature maps $\mathbf{f}_\mathrm{neut} = \mathcal{D}_\mathrm{neut}(\mathbf{Z}_\mathrm{id}; \Phi_\mathrm{neut})$. Supervised by registered neutral 3D scan data during training, $\mathbf{f}_\mathrm{neut}$ encapsulates the subject's base geometry and appearance. This dedicated Neutral Decoder is a key design choice; by explicitly learning to inject these identity-specific features, it promotes better disentanglement of identity from expression, ensures more robust identity preservation during animation, and contributes to more stable training, ultimately leading to sharper, higher-fidelity rendering. Critically, this enables efficient personalization to unseen captures of new identities (Sec.~\ref{sec:personalization}), without requiring explicit neutral 3D scans for those new subjects.
(2) A Guide Mesh Decoder $\mathcal{D}_\mathrm{guide}$ predicts vertex positions $\mathbf{\hat{v}}_p$. Conditioned on both $\mathbf{Z}_\mathrm{id}$ and $\mathbf{Z}_\mathrm{exp}$, this decoder, $\mathcal{D}_\mathrm{guide}(\mathbf{Z}_\mathrm{id}, \mathbf{Z}_\mathrm{exp}; \Phi_\mathrm{guide})$, predicts these as offsets relative to a canonical template mesh, $\mathbf{v}_\mathrm{can}$, which has a fixed topology of 7306 vertices.
(3) The Gaussian Avatar Decoder $\mathcal{D}_\mathrm{ga}$, a CNN-based decoder, $\mathcal{D}_\mathrm{ga}(\mathbf{Z}_\mathrm{id}, \mathbf{Z}_\mathrm{exp}, \mathbf{f}_\mathrm{neut}, V; \Phi_\mathrm{ga})$, predicts the parameters $\{ \delta \mathbf{t}_k, \mathbf{q}_k, \mathbf{s}_k, o_k, \mathbf{c}_k \}$ for the set of 3D Gaussians. The Gaussians are learned on a UV map that is parameterized by the guide mesh topology. $\delta \mathbf{t}_k$ represents predicted offsets from initial Gaussian positions which are initialized on the decoded guide mesh vertices $\mathbf{\hat{v}}_p$. This decoder is conditioned on $\mathbf{Z}_\mathrm{id}$, $\mathbf{Z}_\mathrm{exp}$, the identity features $\mathbf{f}_\mathrm{neut}$ (injected at various network layers), and the camera viewpoint $V$.
The final rendered image is denoted as $I_{p} = \mathcal{R}(\{g_k\})$.
%
\subsubsection{UHAP Training Objective} \label{sec:uhap_loss}
For UHAP training, we utilize the Ava-256 dataset~\cite{martinez2024codec}. This dataset provides multi-view images for 256 subjects and, critically for our loss terms, includes annotations such as: non-rigid mesh tracking for dynamic geometry ($G_\mathrm{exp}$), which yields the ground truth vertices $\mathbf{v}_\mathrm{g}$ maintaining a consistent topology across expressions and subjects; tracked dynamic appearance as UV maps ($T_\mathrm{exp}$); and per-subject neutral scan data ($G_\mathrm{neu}, T_\mathrm{neu}$). The neutral data ($G_\mathrm{neu}, T_\mathrm{neu}$) is derived by averaging the tracked dynamic geometry and texture sequences for each subject.
We jointly optimize all UHAP parameters $\Phi=\bigl(\Phi_{E_{\mathrm{exp}}},\allowbreak\ \Phi_{\mathcal D_{\mathrm{UHAP}}}\bigr)$, where $\Phi_{\mathcal D_{\mathrm{UHAP}}}=\bigl(\Phi_{\mathrm{neut}},\allowbreak\ \Phi_{\mathrm{guide}},\allowbreak\ \Phi_{\mathrm{ga}}\bigr)$.
The overall loss $\mathcal L_{\mathrm{UHAP}}$ is defined as following:
%
\begin{equation} \label{eq:loss_uhap_compact}
\begin{split}
\mathcal{L}_\mathrm{UHAP} = \lambda_\mathrm{rec}\mathcal{L}_\mathrm{rec} + \lambda_\mathrm{neut}\mathcal{L}_\mathrm{neut} + \lambda_\mathrm{KL}\mathcal{L}_\mathrm{KL} + \lambda_\mathrm{geo}\mathcal{L}_\mathrm{geo} \\
+ \lambda_\mathrm{perc}\mathcal{L}_\mathrm{perc} + \lambda_\mathrm{reg\_id}\mathcal{L}_\mathrm{reg\_id} + \lambda_\mathrm{reg\_gauss}\mathcal{L}_\mathrm{reg\_gauss}
\end{split}
\end{equation}
%
Here, $\mathcal{L}_\mathrm{rec}$ is an image reconstruction loss (L1 and SSIM~\cite{wang2004_image-quality}) between the rendering $I_\mathrm{p}$ and ground truth image $I_\mathrm{g}$.
$\mathcal{L}_\mathrm{neut}$ is an L1 loss on the model's reconstruction of the neutral scan data ($G_\mathrm{neu}, T_\mathrm{neu}$) provided by the Ava-256 dataset.
$\mathcal{L}_\mathrm{KL}$ is the KL-divergence for the VAE's expression posterior $q(\mathbf{Z}_\mathrm{exp}|\Delta T_\mathrm{exp}, \Delta G_\mathrm{exp})$ against $\mathcal{N}(0, I)$.
$\mathcal{L}_\mathrm{geo}$ is an L2 loss comparing the predicted guide mesh vertices $\mathbf{\hat{v}}_\mathrm{p}$ to the ground truth tracked vertices $\mathbf{v}_\mathrm{g}$, which are obtained from the non-rigid mesh tracking annotations in the Ava-256 dataset.
$\mathcal{L}_\mathrm{perc}$ is a perceptual loss~\cite{JohnsonAL16} between $I_\mathrm{p}$ and $I_\mathrm{g}$.
$\mathcal{L}_\mathrm{reg\_id}$ is an L1 norm on the identity code $\mathbf{Z}_\mathrm{id}$.
$\mathcal{L}_\mathrm{reg\_gauss}$ includes standard 3D Gaussian regularizations (e.g., for opacity and scale)~\cite{teotiagaussian}.
The $\lambda_{(\cdot)}$ values are hyperparameter weights.
%
%
\subsubsection{Monocular Expression Encoder} \label{sec:mono_encoder}
To effectively personalize our UHAP model to new, unseen captures (videos or static images) and to facilitate image-driven animation, we train a dedicated Monocular Expression Encoder, $E_\mathrm{image}(I_i; \Phi_\mathrm{img})$.
This network's primary role is to map an input image $I_\mathrm{i}$ to an estimated expression code $\mathbf{\hat{Z}}_\mathrm{exp}$ within UHAP's learned latent expression space.
By explaining expression-dependent variations in the input image, $E_\mathrm{image}$ allows the subsequent fine-tuning of $\mathcal{D}_\mathrm{UHAP}$ (Sec.~\ref{sec:personalization}) to focus on capturing the global, identity-specific attributes of the new subject.

$E_\mathrm{image}$ is trained using frontal images derived from our UHAP training data as inputs, with the corresponding ground truth expression codes $\mathbf{Z}_\mathrm{exp}$ (obtained from $E_\mathrm{exp}$ as described in Sec.~\ref{sec:uhap_exp_enc}) serving as targets.
To enhance its generalization capabilities and encourage the learning of identity-agnostic expression features, we employ a data augmentation strategy during the training of $E_\mathrm{image}$.
This involves randomly swapping the identity of the input renderings by leveraging LivePortrait~\cite{guo2024liveportrait} as an effective expression-transfer tool to re-render the same expression on a different identity. The objective $\mathcal{L}_{E_\mathrm{image}}$ combines a squared L2 loss on the predicted latent codes with an L1 reconstruction loss.
This L1 loss measures the difference between renderings produced using the predicted expression code ($\hat{I}_\mathrm{p}$ for images, $\mathbf{\hat{v}_\mathrm{p}}$ for guide mesh vertices) and pseudo-ground truth targets ($\hat{I}_\mathrm{g}, \mathbf{\hat{v}}_\mathrm{g}$):
%
\begin{equation} \label{eq:loss_eimage_final_compact}
\mathcal{L}_{E_\mathrm{image}} = \lambda_\mathrm{latent}||\mathbf{\hat{Z}}_\mathrm{exp} -\mathbf{Z}_\mathrm{exp}||_2 + \lambda_\mathrm{recon} (||\hat{I}_\mathrm{p} - \hat{I}_\mathrm{g}||_1 + ||\mathbf{\hat{v}_\mathrm{p}} - \mathbf{\hat{v}}_\mathrm{g}||_1)
\end{equation}
%
Here, $\hat{I}_\mathrm{p}$ and $\mathbf{\hat{v}_\mathrm{p}}$ are generated using $\mathcal{D}_\mathrm{UHAP}(\mathbf{Z}_\mathrm{id}, \mathbf{\hat{Z}}_\mathrm{exp})$, while $\hat{I}_\mathrm{g}$ and $\mathbf{\hat{v}}_\mathrm{g}$ are the pseudo-ground truth targets derived from UHAP training data, as depicted in Fig.~\ref{fig:overview} (middle).
%
\begin{figure}[h!]
\centering
\includegraphics[width=0.89\linewidth]{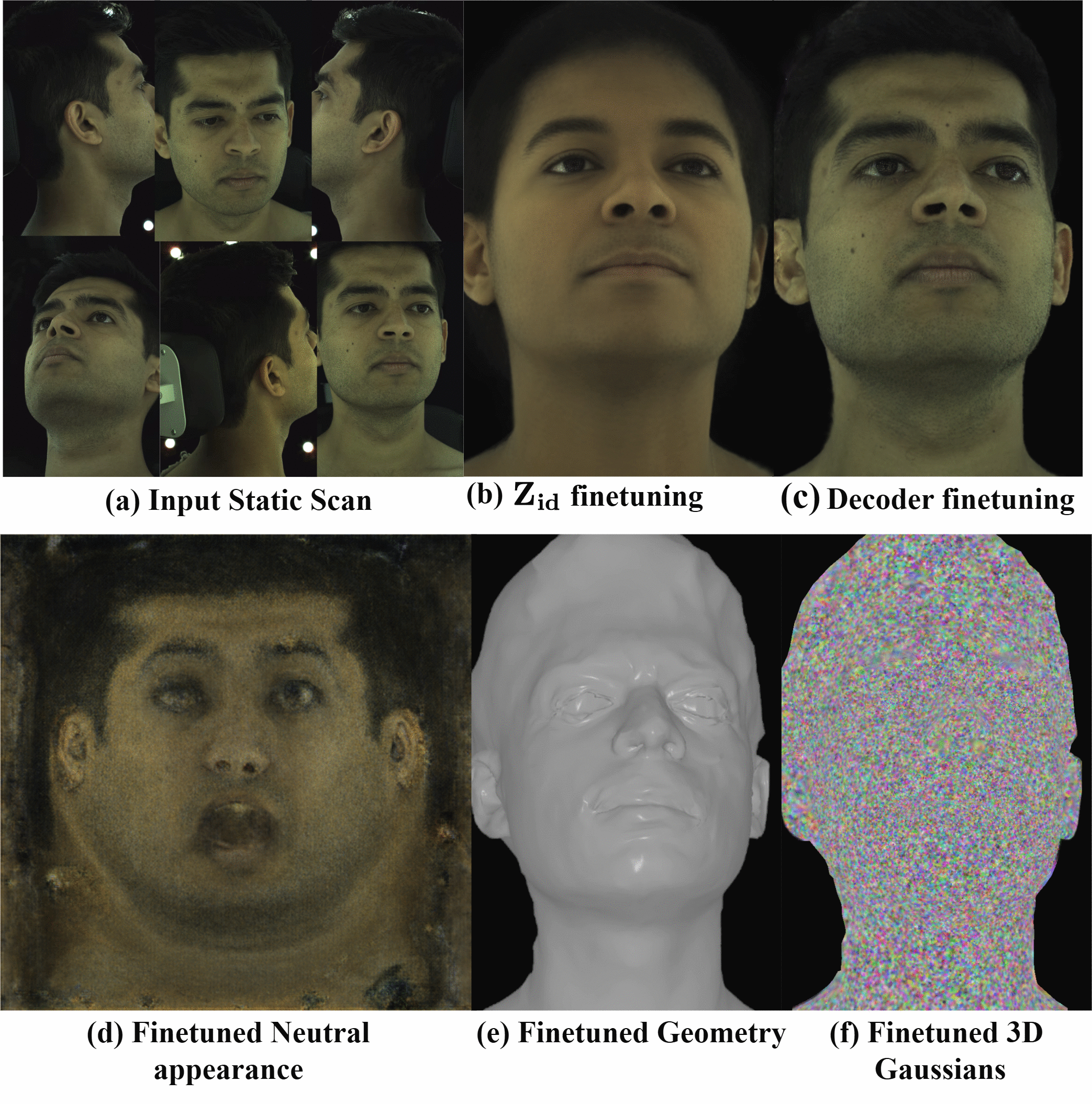} 
\caption{
Personalization pipeline stages: 
(a) Input static scan. 
(b) Result after $\mathbf{Z}_\mathrm{id}$ fine-tuning. 
(c) Result after $\mathcal{D}_\mathrm{UHAP}$ decoder fine-tuning. 
Process yields (d) finetuned neutral appearance, (e) geometry, and (f) 3D Gaussians.
}
\label{fig:personalization}
\end{figure}
\subsection{Audio-Driven Avatar Synthesis} \label{sec:audio_synthesis}
To animate our Universal Head Avatar Prior (UHAP) from speech (Figure~\ref{fig:overview}, Right), we generate sequences of its rich expression codes, $\mathbf{Z_{exp}}$. These codes are designed to holistically modulate the entire facial state. Generating full-face expressions directly from audio, which primarily correlates with lip movements, is a key challenge. Our core audio-to-expression generator is a diffusion probabilistic model (DDPM)~\cite{NEURIPS2020_4c5bcfec}, $\mathcal{G}_{\theta}$. For its backbone, we use the Transformer-based model as proposed in ~\cite{ng2024audiophotorealembodimentsynthesizing}. While the framework in \cite{ng2024audiophotorealembodimentsynthesizing} is effective for generating expressive outputs, it was originally applied to predict person-specific codes. In contrast, our $\mathcal{G}_{\theta}$ is trained to synthesize sequences within our person-agnostic UHAP expression space $\mathbf{Z_{exp}}$. Furthermore, our approach differs from other diffusion-based models like FaceTalk~\cite{aneja2023facetalk}, which, though also using a Transformer architecture, primarily predicts latent codes for geometry-only parametric models. Our $\mathbf{Z_{exp}}$ latents, conversely, drive both the geometry and the appearance-related facial expression dynamics of UHAP. The DDPM $\mathcal{G}_{\theta}$ is conditioned on several inputs: audio features $\mathbf{A}^{1:N}$, predicted lip vertices $\mathbf{L}_{v}$, the noisy expression codes $\mathbf{Z^{t}_{exp}}$, and the diffusion timestep $t$. The audio features $\mathbf{A}^{1:N}$ are extracted from the input waveform by a Wav2Vec-based encoder~\cite{baevski2020wav2vec20frameworkselfsupervised} ($E_{audio}$ in Figure~\ref{fig:overview}). The lip vertices $\mathbf{L}_{v}$ are predicted by a dedicated Audio-to-lip Module ($M_{a2l}$ in Figure~\ref{fig:overview}) from $\mathbf{A}^{1:N}$ to provide strong local synchronization cues. The Audio-to-lip module uses the Wav2Vec encoder \cite{baevski2020wav2vec20frameworkselfsupervised} and a pretrained, lightweight transformer to predict $338$ lip vertices directly from audio. These vertices provide explicit local conditioning for our diffusion model. The Transformer architecture~\cite{vaswani2023attentionneed} within $\mathcal{G}_{\theta}$ utilizes self-attention, cross-attention to fuse these conditioning signals, and FiLM layers~\cite{perez2018film} for the timestep embedding.

$\mathcal{G}_{\theta}$ is trained to predict the noise $\boldsymbol{\epsilon}$ added to the clean expression codes $\mathbf{Z^{0}_{exp}}$, using the standard DDPM objective:
\begin{equation} \label{eq:simple_loss_corrected_v4}
\mathcal{L}_{diff} = \mathbb{E}_{\mathbf{Z_{exp}^{0}}, \mathbf{A}, \mathbf{L}_{v}, \boldsymbol{\epsilon}, t} \left[ \left\| \boldsymbol{\epsilon} - \mathcal{G}_\theta\left( \sqrt{\bar{\alpha}_t}\mathbf{Z_{exp}^{0}} + \sqrt{1-\bar{\alpha}_t}\boldsymbol{\epsilon}, t, \mathbf{A}, \mathbf{L}_{v} \right) \right\|_2^2 \right]
\end{equation}
where $\boldsymbol{\epsilon} \sim \mathcal{N}(0,I)$ and $\bar{\alpha}_t$ is from the noise schedule. Training employs paired audio segments and $\mathbf{Z_{exp}}$ codes from the Multiface dataset~\cite{wuu2023multifacedatasetneuralface}, where $\mathbf{Z_{exp}}$ codes are obtained via our UHAP's $E_{exp}$ (Sec.~\ref{sec:uhap_exp_enc}). During inference, the denoised sequence $\mathbf{\hat{Z}^{0}_{exp}}$, with a target identity code $\mathbf{Z_{id}}$, drives the frozen UHAP decoder $\mathcal{D}_{UHAP}$.

%
\subsection{Personalization for New Identities} \label{sec:personalization}
Our UHAP can be personalized to new identities using various input data, including short dynamic captures of the new subject, or a static multi-view capture.
A key advantage of our personalization approach is its efficiency and minimal data prerequisites: for the input data, we only require the rigid head pose and do not necessitate prior non-rigid 3D tracking or complex geometric registration of the subject. To adapt UHAP to a new identity, for instance from a static capture, we perform a two-stage fine-tuning process.
This entire process takes approximately 20 minutes in total on a single NVIDIA A40 GPU.
When adapting UHAP to a new identity from a static scan, we pass a frontal image from the scan to our pre-trained Monocular Expression Encoder $E_{image}$ (Sec.~\ref{sec:mono_encoder}) to obtain the corresponding expression code, $\mathbf{Z_{exp}}$. This specific expression code $\mathbf{Z_{exp}}$ is then held constant throughout the subsequent two-stage fine-tuning procedure. If personalizing from a short dynamic video, $E_{image}$ would provide per-frame expression codes.
First, the identity code $\mathbf{Z}_\mathrm{id}$ is optimized for $\sim$2k iterations (Fig.~\ref{fig:personalization}b).
Second, the UHAP decoder $\mathcal{D}_\mathrm{UHAP}$ is fine-tuned for $\sim$2k iterations (Fig.~\ref{fig:personalization}c) using the fitting loss $\mathcal{L}_\mathrm{fit}$:
%
\begin{equation}
\label{eq:loss_fit}
\mathcal{L}_\mathrm{fit} = \alpha_1 \mathcal{L}_\mathrm{photo} + \alpha_2 \mathcal{L}_\mathrm{laplacian} + \alpha_3 \mathcal{L}_\mathrm{offset} + \alpha_4 \mathcal{L}_\mathrm{scale}
\end{equation}
%
where $\mathcal{L}_\mathrm{photo}$ is an $\mathcal{L}_1$ photometric loss between the rendered image and the input scan; $\mathcal{L}_\mathrm{laplacian}$ regularizes the smoothness of the guide mesh vertices ($\mathbf{v}_t$); and $\mathcal{L}_\mathrm{offset}$ and $\mathcal{L}_\mathrm{scale}$ are $\mathcal{L}_1$ norms applied to the predicted Gaussian positional offsets and scales, respectively.
The coefficients $\alpha_i$ are hyperparameter weights balancing these terms.
This two-stage process yields the subject's personalized neutral appearance (Fig.~\ref{fig:personalization}d), geometry (e), and the set of 3D Gaussians (f) that constitute the fine-tuned 3D Gaussian Avatar.

%
%
%
%
\section{Experiments} \label{sec:experiments}
\begin{figure*}[htb]
\centering
\includegraphics[width=0.8\textwidth]{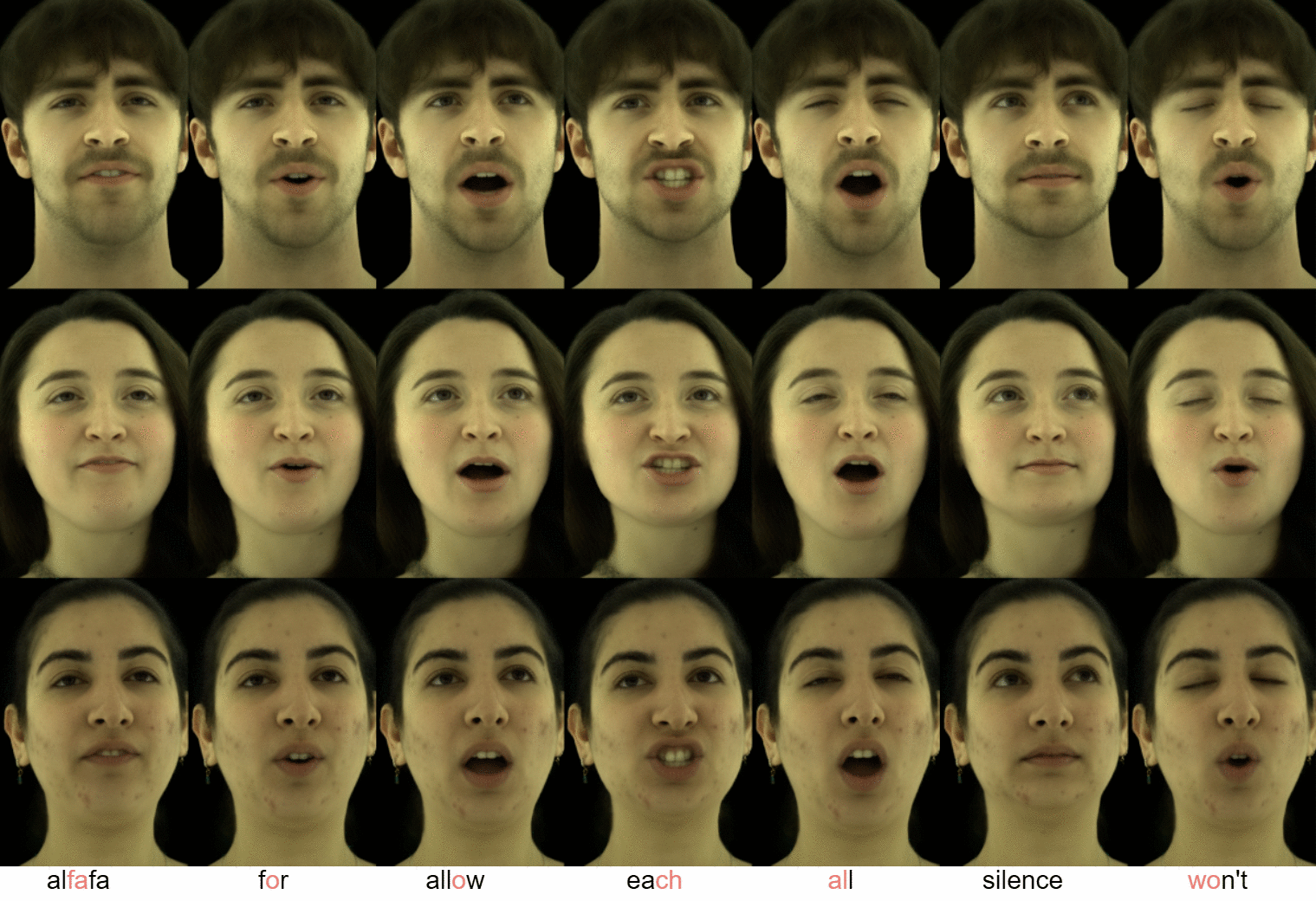} 
\caption{Audio-driven synthesis results for three UHAP model identities with corresponding audio prompts.}
\label{fig:qualitative_audio_diverse_identities}
\end{figure*}
\begin{figure*}[htb]
\centering
\includegraphics[width=\textwidth]{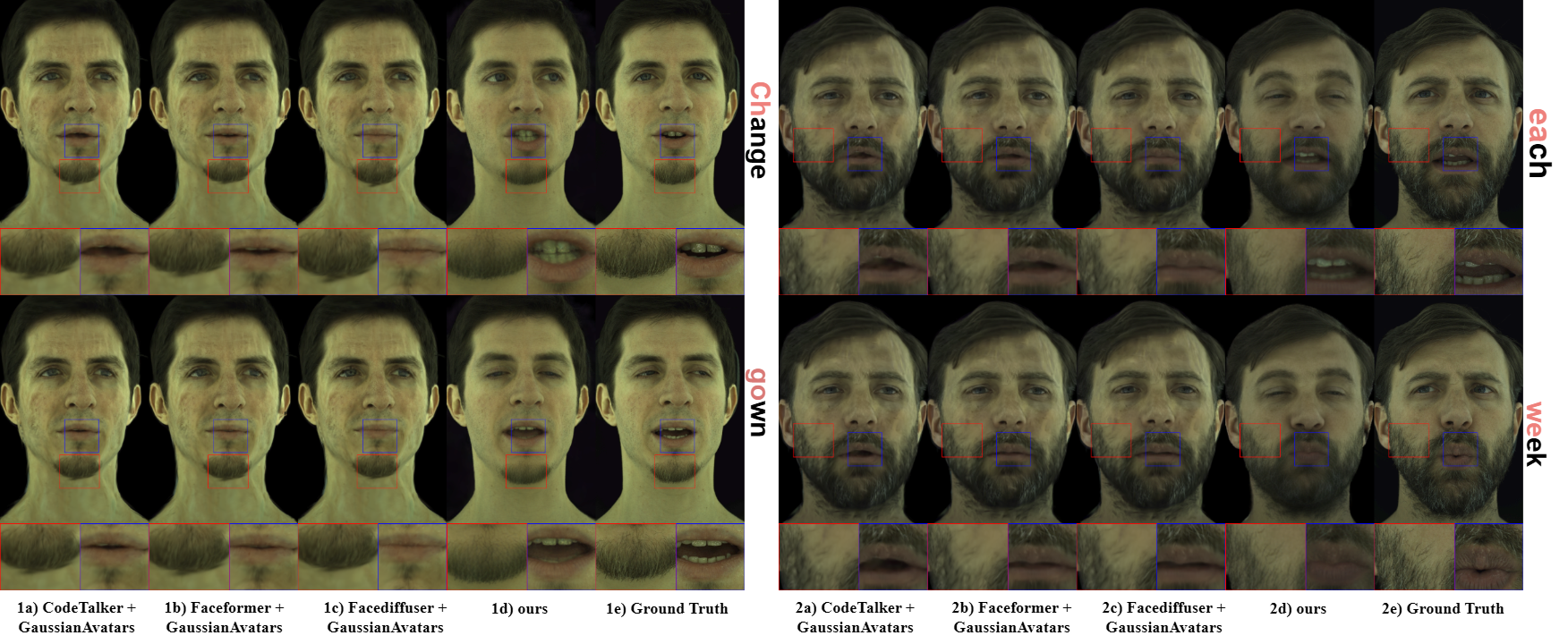} 
\caption{Qualitative comparison with SOTA methods (CodeTalker+GA, Faceformer+GA, FaceDiffuser+GA, Ours) and Ground Truth for specified audio segments. GA denotes GaussianAvatars augmentation. }
\label{fig:sota_comparison_visual}
\end{figure*}
In this section, we first outline the datasets used for training our universal prior and the audio-driven synthesis model, along with key implementation details (Sec.~\ref{sec:datasets}).
We then present qualitative results that demonstrate the capabilities of our method in generating audio-driven, diverse, and expressive animations (Sec.~\ref{sec:qualitative_results}).
Subsequently, we provide quantitative and qualitative comparisons against audio-driven (geometric) facial animation methods (Sec.~\ref{sec:comparisons}).
Finally, we conduct ablation studies (Sec.~\ref{sec:ablation_studies}) to validate the impact of our key components and design choices within our proposed framework, such as the role of neutral features, the pretraining of our monocular encoder, and the amount of data needed for personalization.
%
%
\subsection{Datasets and Implementation Details} \label{sec:datasets}
%
%
\par \noindent\textbf{Training Data.}
Our Universal Head Avatar Prior (UHAP) is trained using 230 distinct identities from the Ava-256 dataset~\cite{martinez2024codec}.
This dataset provides multi-view dynamic video recordings and registered neutral 3D scans for each subject \emph{but it contains no audio}. 
The large number of identities in Ava-256 is crucial for learning a robust and generalizable prior (UHAP) over identity and expression.
For training the audio-to-expression synthesis model, we utilize the Multiface dataset~\cite{wuu2023multifacedatasetneuralface}. 
Multiface provides synchronized multi-view video data of subjects uttering a combined 650 sentences, offering rich audio-visual correspondence, though with a more limited number of identities compared to Ava-256. 
All identities from Multiface are unseen during the training of our UHAP prior. 
Multiface also provides tracked dynamic geometry ($G_\mathrm{exp}$), dynamic appearance UV maps ($T_\mathrm{exp}$), and corresponding neutral data ($G_\mathrm{neu}, T_\mathrm{neu}$) for its subjects. 
We leverage our pre-trained UHAP and its associated expression encoder (Sec.~\mbox{\ref{sec:uhap_exp_enc}}) to process these $T_\mathrm{exp}, G_\mathrm{exp}$ sequences from the Multiface dataset, mapping them into our subject-agnostic expression space to obtain $\mathbf{Z_{exp}}$ codes. 
This allows us to create the synchronized audio-feature-to-$\mathbf{Z_{exp}}$ pairs necessary for training our audio-driven model.
Data from $10$ identities from Multiface dataset are used for training this audio model. 
Three Multiface dataset identities, entirely unseen by both the UHAP model and the audio model during their respective training phases, are held out exclusively for testing and evaluating the audio-visual performance of our complete pipeline.
%
%
\par \noindent\textbf{Baseline Setup.}
Our method's capability to personalize to new subjects is versatile, accommodating inputs such as static captures or dynamic videos (as detailed in Sec.~\ref{sec:personalization}). 
For the quantitative and qualitative comparisons against state-of-the-art methods requiring personalization (Sec.~\ref{sec:comparisons}), we use a consistent setup for, both, our model and the baselines. 
Specifically, for new identities from the Multiface test set, our UHAP model is fine-tuned using approximately 500 frames (or ~5 sentences) per camera from 12 views.
We highlight that there is no prior work that is generalizable from speech input and enables photoreal renderings.
Instead, we compare against state-of-the-art geometry-based methods, i.e., Faceformer~\cite{dummy_faceformer_2022}, CodeTalker~\cite{codetalkerspeechdriven3dfacialanimationw}, and FaceDiffuser~\cite{facediffuserspeechdriven3dfacialanimatio}.
Their output consists of animated mesh sequences in FLAME topology, which we further augmented for photorealistic rendering for fair comparison.
This is achieved by training person-specific GaussianAvatars~\cite{dummy_3dgs_avatar_works} for each test identity.
Crucially, these GaussianAvatars are also trained using the identical data setup as our personalization stage leverages.
This ensures a fair and direct comparison in terms of the input data provided for achieving photorealistic results.
%

%
\subsection{Qualitative Results} 
\begin{figure}[h!]
\centering
\includegraphics[width=0.95\linewidth]{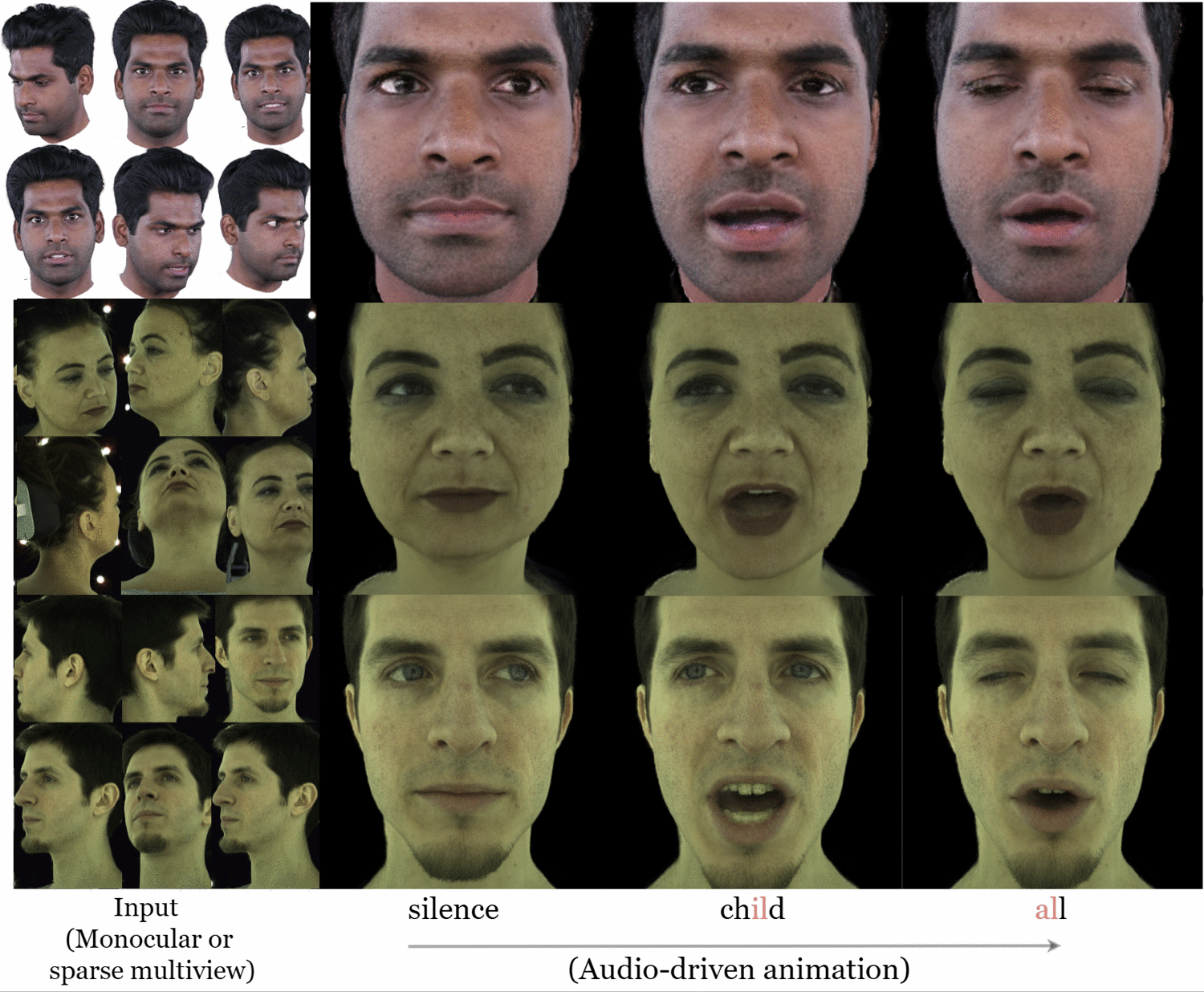} 
\caption{Personalization from sparse inputs (left column) and resulting audio-driven animation (right columns) for three subjects at a novel viewpoint.}
\label{fig:personalization_animation_showcase}
\end{figure}
\begin{figure}[h!]
\centering
\includegraphics[width=0.85\linewidth]{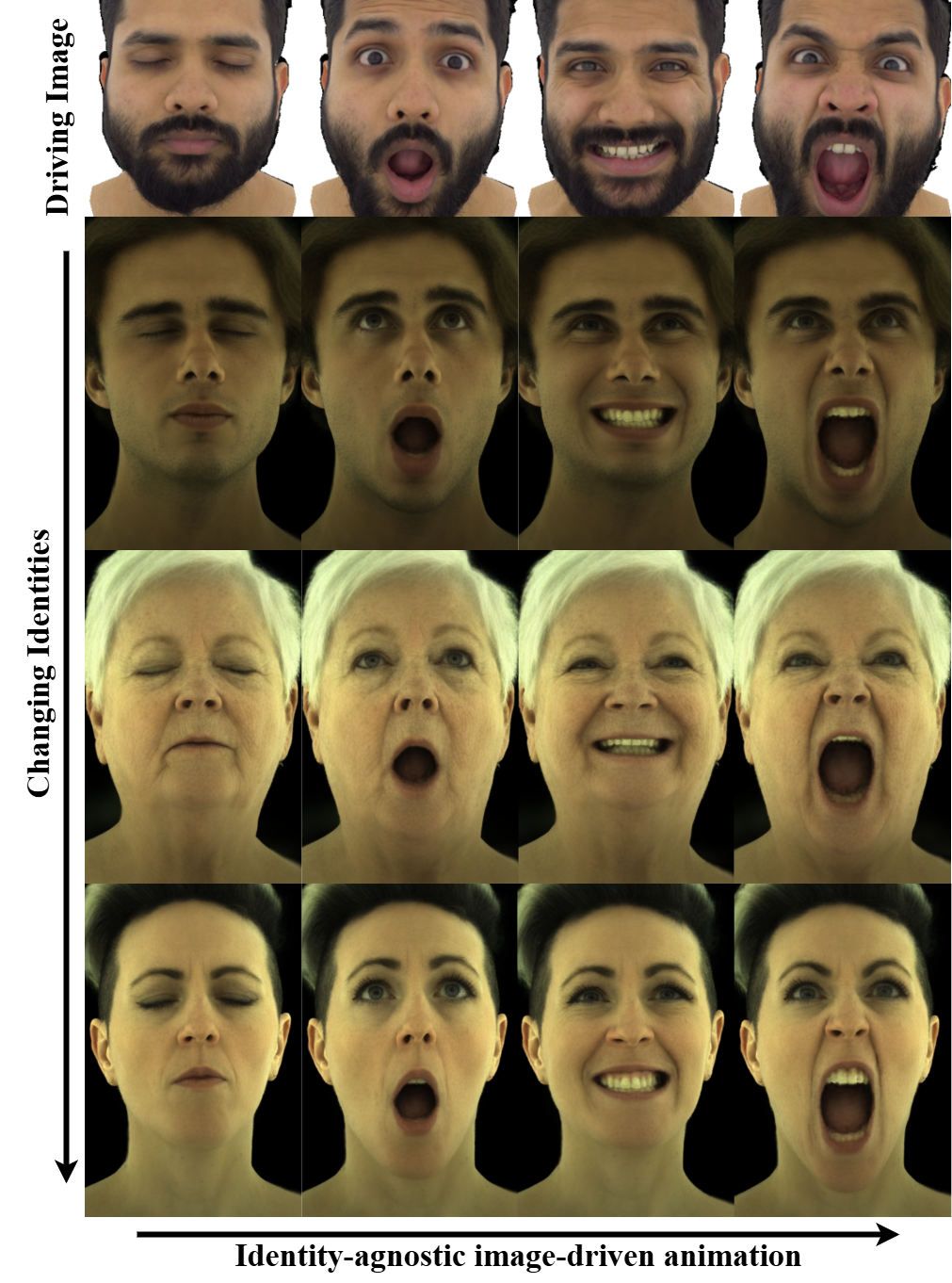} 
\caption{Image-driven animation: Driving image sequence (top) and animated target identities (rows below).}
\label{fig:image_driven_animation_example}
\end{figure}

\label{sec:qualitative_results}
We first show the general qualitative performance of our audio-driven avatar synthesis method.
Fig.~\ref{fig:qualitative_audio_diverse_identities} demonstrates the capability of our method to generate expressive audio-driven animations for three distinct synthesized identities. 
The sequences highlight accurate lip synchronization corresponding to the provided audio prompts, accompanied by natural-looking facial dynamics and varied expressions indicative of the spoken content.
Fig.~\ref{fig:personalization_animation_showcase} illustrates the versatility and effectiveness of our personalization process across various input conditions. 
We show adaptation to new subjects from: 
a monocular video capture from the INSTA dataset \cite{zielonka2022insta}  (top row),  multiview captures (8 input views) from Multiface Dataset \cite{wuu2023multifacedatasetneuralface} (middle row), and (bottom row). We highlight that all these datasets are not part of the UHAP training.
In each case, the input data (leftmost column) is used to personalize UHAP, and the subsequent audio-driven animations (right columns) demonstrate that the unique identities are well-captured and then faithfully animated with coherent speech motions.
Beyond audio-driven synthesis, Fig.~\ref{fig:image_driven_animation_example} underscores the versatility of our learned expression space through image-driven animation. 
Here, expressions from a source driving sequence (top row) are successfully transferred to multiple distinct target identities (rows below), demonstrating accurate expression re-targeting while consistently maintaining the unique appearance and characteristics of each target avatar.
This highlights the successful latent disentanglement of the identity and expression latent space. We also provide additional qualitative results on subjects from the HQ3DAvatar \cite{teotia2023hq3davatar} and Renderme-360  \cite{pan2024renderme} datasets in the supplementary material.
%
%

\subsection{Comparisons with State-of-the-Art Methods}

%
\label{sec:comparisons}
\begin{table}[t]
\small
  \centering
  \caption{Quantitative comparison with SOTA audio-driven avatar methods on the held-out audio and universal model test subjects.
  Metrics are averaged across all frames and identities.}
  \label{tab:quantitative_comparison_metrics}
  \sisetup{table-align-text-post=false} 
  \begin{tabular}{@{}lS[table-format=2.4]S[table-format=1.4]S[table-format=1.4]S[table-format=1.1]@{}}
    \toprule
    {Method} & {PSNR $\uparrow$} & {LPIPS $\downarrow$} & {SSIM $\uparrow$} & {LSE-D $\downarrow$} \\
    \midrule
    CodeTalker   & 26.23 & 0.37 & 0.6518 & 8.30 \\
    FaceFormer   & 25.93 & 0.38 & 0.6475 & 9.32 \\
    FaceDiffuser & 26.32 & 0.43 & 0.6832 & 8.88 \\
    \midrule
    \textbf{Ours}    & \textbf{27.37} & \textbf{0.29} & \textbf{0.7293} & \textbf{6.32} \\
    \bottomrule
  \end{tabular}
\end{table}
%
We conduct a comparative evaluation of our method against several recent state-of-the-art audio-driven facial animation techniques: Faceformer~\cite{dummy_faceformer_2022}, CodeTalker~\cite{codetalkerspeechdriven3dfacialanimationw}, and FaceDiffuser~\cite{facediffuserspeechdriven3dfacialanimatio}.
As these methods primarily focus on generating 3D mesh deformations, their outputs are rendered using personalized GaussianAvatars~\cite{dummy_3dgs_avatar_works} to enable a fair photorealistic comparison.  We evaluate our method on held-out speakers/subjects from the Multiface dataset \cite{wuu2023multifacedatasetneuralface}.
%
%
%
\par \noindent\textbf{Quantitative Comparison.}
Tab.~\ref{tab:quantitative_comparison_metrics} summarizes the quantitative results on the held-out Multiface test identities.
Our method achieves superior performance across standard image reconstruction metrics, including higher PSNR and lower L1 and LPIPS \cite{zhang2018perceptual} scores, which indicates enhanced image fidelity and perceptual quality.
Furthermore, our approach demonstrates improved audio-visual synchronization, as reflected by a better (lower) LSE-D score \cite{Chung16a}. 
Since these metrics test the end-to-end performance from audio to image quality, these results confirm the combined benefits of our contributions that more directly link audio input and avatar rendering.
%
%
\par \noindent\textbf{Qualitative Comparison.}
Fig.~\ref{fig:sota_comparison_visual} provides a side-by-side visual comparison against state-of-the-art methods, personalized on the same Multiface test subjects, and ground truth for specified audio segments,
 shown from a held-out novel viewpoint.
%
%
Our method consistently produces results with higher fidelity details in terms of appearance and geometry. 
For instance, in subjects with facial hair, our approach renders a sharper beard that deforms naturally and coherently with speech-induced jaw and cheek movements, a detail which prior method can typically not preserve resulting in smoothed out renderings that lack photorealism.
Furthermore, our model generates a significantly sharper and more realistic mouth interior, contributing to more natural expressions during speech.
This, combined with more precise mouth articulation (e.g., for words like ``change'' and ``bride'') and subtle eye movements, leads to better visual lip synchronization and overall fidelity to the ground truth, which is visibly higher compared to the baselines.
This visual superiority can be attributed to our model's ability to directly synthesize these fine-grained appearance attributes coherently with geometric deformations, all driven by the audio-driven latent expression codes.
%
%

\subsection{Ablation Studies} 
\label{sec:ablation_studies}
\begin{figure}[h!]
\centering
\includegraphics[width=0.88\linewidth]{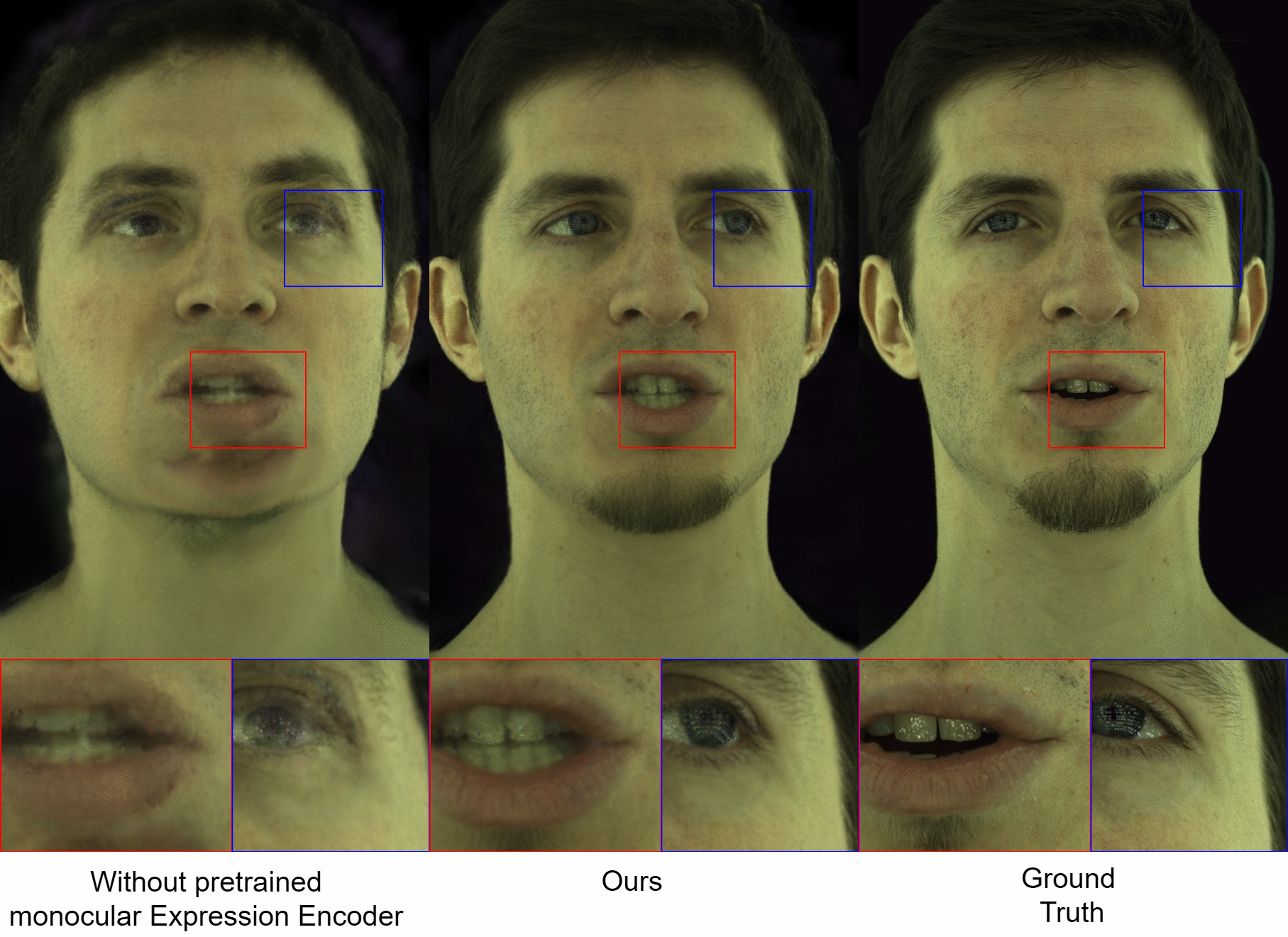} 
\caption{Ablation on monocular encoder ($E_{image}$) training: Encoder trained during fitting (left), Ours (pretrained encoder, center), Ground Truth (right).}
\label{fig:ablation_monocular_encoder}
\end{figure}

\begin{figure}[h!]
\centering
\includegraphics[width=0.85\linewidth]{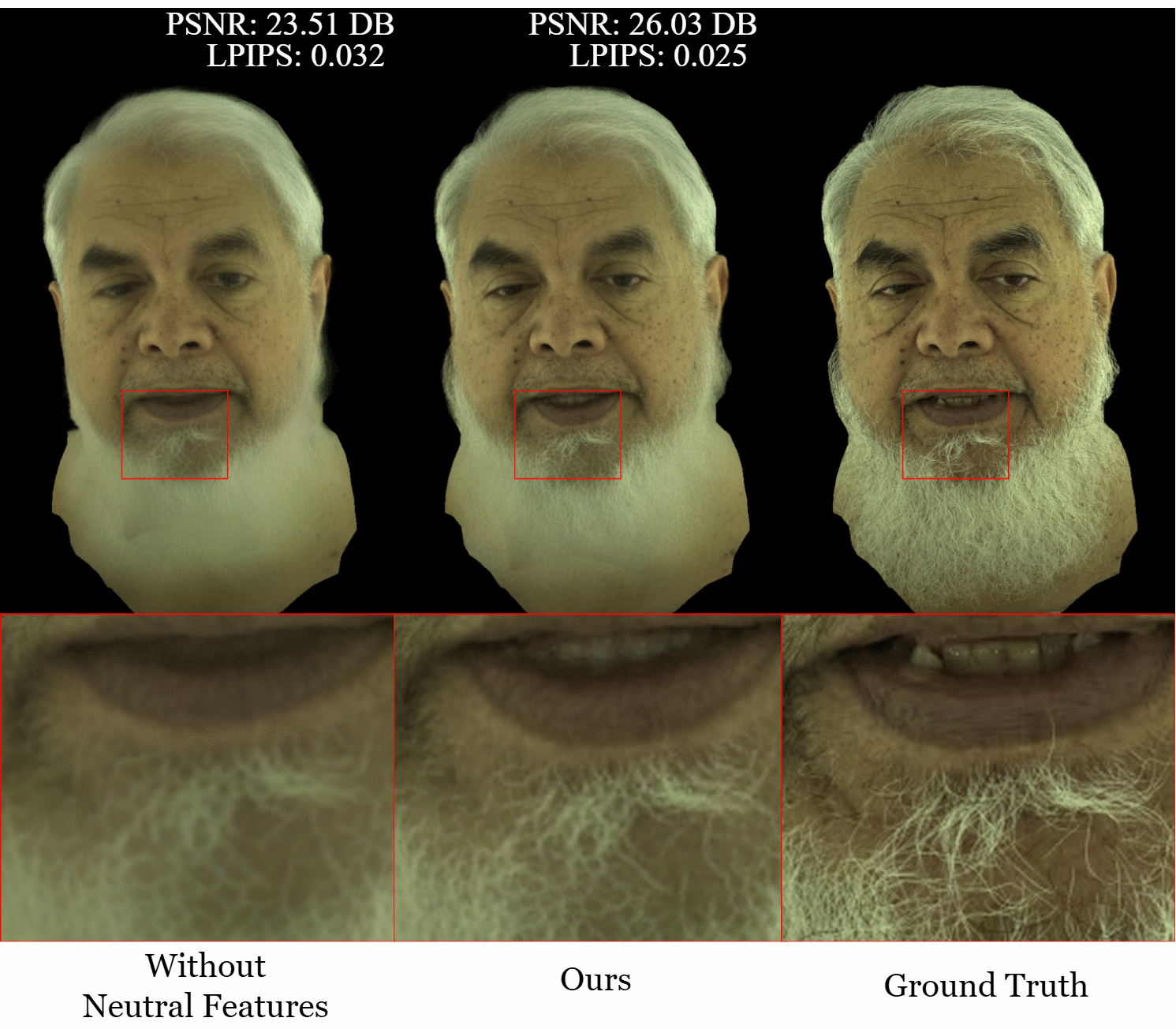} 
\caption{Ablation on neutral features ($f_{neut}$): Without Neutral Features (left), Ours (center), Ground Truth (right), with PSNR/LPIPS metrics.}
\label{fig:ablation_neutral_features}
\end{figure}
To validate the contributions of individual components and design choices within our framework, we perform several ablation studies.
%
%
\par \noindent\textbf{Impact of Neutral Features in UHAP.}
Fig.~\ref{fig:ablation_neutral_features} evaluates the importance of incorporating identity-specific neutral features ($f_\mathrm{neut}$) during UHAP training.
The visual comparison shows renderings with our full model, without neutral features, and the Ground Truth, alongside quantitative metrics.
Removing these neutral feature inputs results in a discernible degradation in rendering quality and the precision of identity preservation.
This highlights the critical role of these learned neutral characteristics in achieving high-fidelity personalization with our UHAP.
%
%
\par \noindent\textbf{Role of Pretrained Monocular Expression Encoder.}
The significance of employing a pretrained monocular expression encoder ($E_\mathrm{image}$), as opposed to training it from scratch during subject-specific fine-tuning, is demonstrated in Fig.~\ref{fig:ablation_monocular_encoder}.
The left panel shows results when the encoder is trained during fitting, the center panel shows our approach with a pretrained encoder, and the right panel shows ground truth.
When the expression encoder is trained concurrently with subject fine-tuning, it tends to learn a mapping that entangles expression with the specific subject's appearance and geometry.
This causes a mismatch when expression codes from our universal audio model, which expects the original, disentangled latent space semantics, are fed into this subject-adapted decoder, leading to distorted expressions and incorrect appearance.
Our proposed approach, which utilizes the pretrained encoder designed to isolate true expression variations, maintains compatibility with the audio model's output, ensuring faithful synthesis.

\section{Conclusion} \label{sec:conclusion}
We have introduced a novel framework for the audio-driven synthesis of universal, photorealistic 3D Gaussian head avatars. Our Universal Head Avatar Prior (UHAP), learns a rich expression latent space that holistically controls both detailed geometry and dynamic appearance. Combined with an efficient personalization strategy adaptable to sparse inputs and an audio-to-expression diffusion model, our approach generates high-fidelity animations. These animations demonstrate accurate lip synchronization and nuanced facial dynamics such as eye-gaze shifts, all generalizing across diverse identities and sparse, monocular capture settings.
\begin{figure}[h!]
\centering
\includegraphics[width=0.92\linewidth]{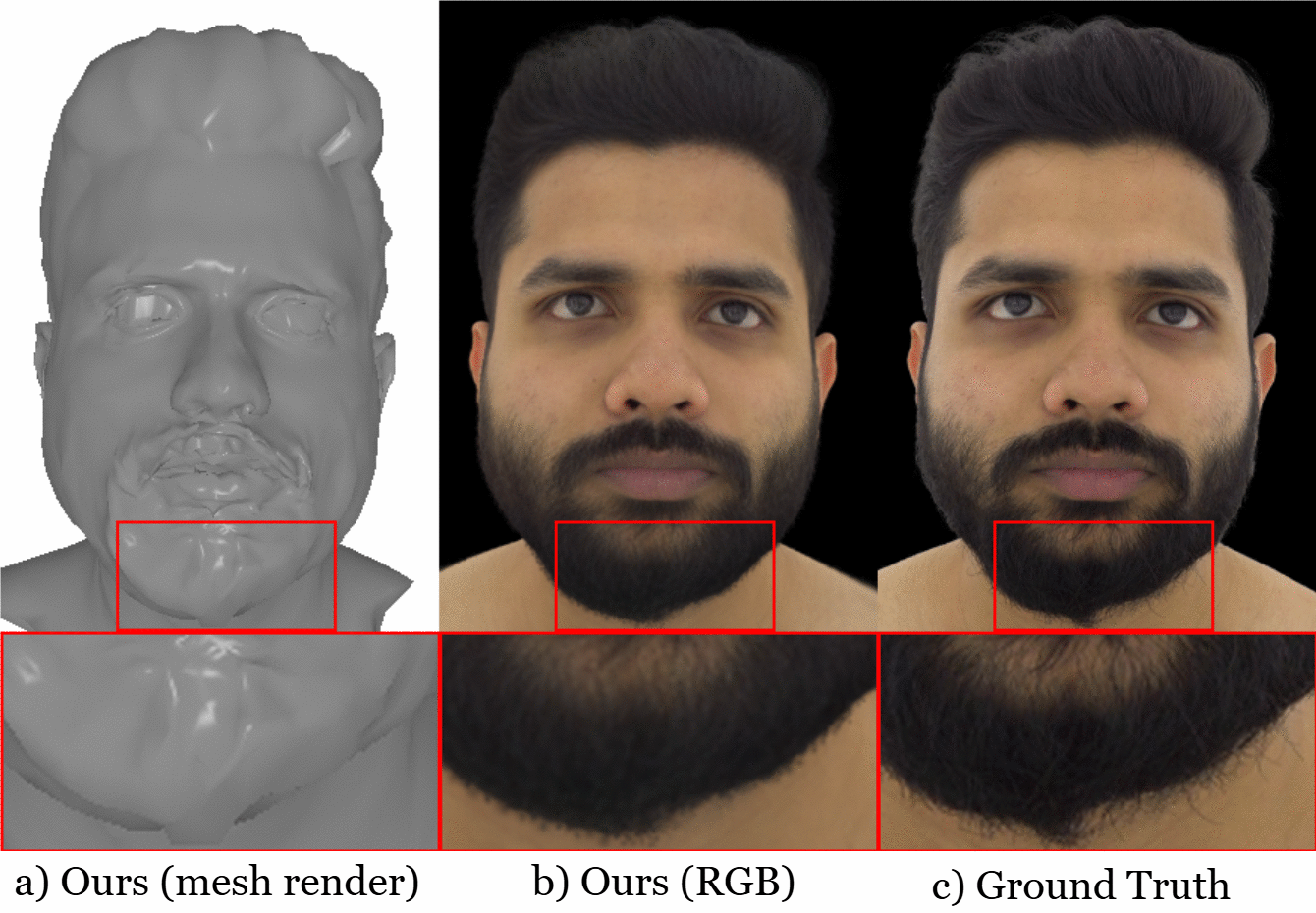} 
\caption{Our method struggles with fine regions such as beards, where our model's geometry-fitting (a) averages out fine details, leading to less accurate reproduction (b) compared to ground truth (c).}
\label{fig:limitaion}
\end{figure}
\par \noindent\textbf{Limitations.}
Despite promising results, our method has limitations. While synthesized upper-head expressions generally align well with speech-driven mouth motion, the current gaze behavior can sometimes appear unnatural, potentially reflecting the script-reading nature of the audio training data. Training on conversational audio-visual data and adding explicit gaze control are potential avenues for future works to overcome this limitation. Furthermore, although our model reconstructs fine details like static hair strands, it struggles with elements that exhibit complex, independent motion relative to the skin surface, such as beards as shown in Fig.~\ref{fig:limitaion}, which the current representation may not perfectly register. This limitation can be overcome by leveraging strand-based representation for facial hair \cite{dinsey_beard}. While our appearance model can render avatars in real-time ($50$ FPS on a single NVIDIA A40 series GPU), the audio-to-latent module does not decode expression codes in real-time due to the iterative nature of the denoising process of the diffusion model. Future work could explore diffusion models optimized for faster inference to enable real-time expression code decoding while maintaining quality. Finally, our UHAP is trained on high-quality studio data with the same lighting conditions across subjects. Robustness to in-the-wild captures (e.g., mobile phone recordings under uncontrolled lighting) is still limited. Extending the UHAP training corpus to include light-stage data will increase robustness to such capture conditions.

\par \noindent\textbf{Future Work.}
Future work will aim to address these limitations and further enhance our system's capabilities. We plan to investigate methods for learning more natural and interactive gaze behaviors, perhaps by incorporating data from unscripted conversational videos or by enabling explicit gaze control. A significant avenue for future development involves leveraging our monocular expression encoder ($E_{image}$) by using it as an inference tool to collect expression codes on large-scale, diverse in-the-wild audio-visual datasets. This could enable the explicit modeling and audio-driven synthesis of a broader spectrum of nuanced human emotions, thereby enriching avatar expressiveness and realism.

\section{Acknowledgements}
We would like to thank Flawless AI for financially supporting Kartik Teotia for this work. 

\bibliographystyle{ACM-Reference-Format}
\bibliography{references}


\begin{thebibliography}{66}


\ifx \showCODEN    \undefined \def \showCODEN     #1{\unskip}     \fi
\ifx \showISBNx    \undefined \def \showISBNx     #1{\unskip}     \fi
\ifx \showISBNxiii \undefined \def \showISBNxiii  #1{\unskip}     \fi
\ifx \showISSN     \undefined \def \showISSN      #1{\unskip}     \fi
\ifx \showLCCN     \undefined \def \showLCCN      #1{\unskip}     \fi
\ifx \shownote     \undefined \def \shownote      #1{#1}          \fi
\ifx \showarticletitle \undefined \def \showarticletitle #1{#1}   \fi
\ifx \showURL      \undefined \def \showURL       {\relax}        \fi
\providecommand\bibfield[2]{#2}
\providecommand\bibinfo[2]{#2}
\providecommand\natexlab[1]{#1}
\providecommand\showeprint[2][]{arXiv:#2}

\bibitem[Aneja et~al\mbox{.}(2024a)]%
        {aneja2024gaussianspeech}
\bibfield{author}{\bibinfo{person}{Shivangi Aneja}, \bibinfo{person}{Artem
  Sevastopolsky}, \bibinfo{person}{Tobias Kirschstein}, \bibinfo{person}{Justus
  Thies}, \bibinfo{person}{Angela Dai}, {and} \bibinfo{person}{Matthias
  Nießner}.} \bibinfo{year}{2024}\natexlab{a}.
\newblock \bibinfo{title}{GaussianSpeech: Audio-Driven Gaussian Avatars}.
\newblock
\showeprint[arxiv]{2411.18675}~[cs.CV]
\urldef\tempurl%
\url{https://arxiv.org/abs/2411.18675}
\showURL{%
\tempurl}


\bibitem[Aneja et~al\mbox{.}(2024b)]%
        {aneja2023facetalk}
\bibfield{author}{\bibinfo{person}{Shivangi Aneja}, \bibinfo{person}{Justus
  Thies}, \bibinfo{person}{Angela Dai}, {and} \bibinfo{person}{Matthias
  Nießner}.} \bibinfo{year}{2024}\natexlab{b}.
\newblock \showarticletitle{FaceTalk: Audio-Driven Motion Diffusion for Neural
  Parametric Head Models}. In \bibinfo{booktitle}{\emph{Proc. IEEE Conf. on
  Computer Vision and Pattern Recognition (CVPR)}}.
\newblock


\bibitem[Aneja et~al\mbox{.}(2025)]%
        {aneja2025scaffoldavatar}
\bibfield{author}{\bibinfo{person}{Shivangi Aneja}, \bibinfo{person}{Sebastian
  Weiss}, \bibinfo{person}{Irene Baeza}, \bibinfo{person}{Prashanth Chandran},
  \bibinfo{person}{Gaspard Zoss}, \bibinfo{person}{Matthias Nießner}, {and}
  \bibinfo{person}{Derek Bradley}.} \bibinfo{year}{2025}\natexlab{}.
\newblock \bibinfo{title}{ScaffoldAvatar: High-Fidelity Gaussian Avatars with
  Patch Expressions}.
\newblock
\showeprint[arxiv]{2507.10542}~[cs.GR]
\urldef\tempurl%
\url{https://arxiv.org/abs/2507.10542}
\showURL{%
\tempurl}


\bibitem[Aylagas et~al\mbox{.}(2022)]%
        {voice2faceaudiodrivenfacialandtongueriga}
\bibfield{author}{\bibinfo{person}{Monica~Villanueva Aylagas},
  \bibinfo{person}{Hector~Anadon Leon}, \bibinfo{person}{Mattias Teye}, {and}
  \bibinfo{person}{Konrad Tollmar}.} \bibinfo{year}{2022}\natexlab{}.
\newblock \bibinfo{title}{Voice2face: Audio-driven facial and tongue rig
  animations with cvaes}.
\newblock


\bibitem[Baevski et~al\mbox{.}(2020)]%
        {baevski2020wav2vec20frameworkselfsupervised}
\bibfield{author}{\bibinfo{person}{Alexei Baevski}, \bibinfo{person}{Henry
  Zhou}, \bibinfo{person}{Abdelrahman Mohamed}, {and} \bibinfo{person}{Michael
  Auli}.} \bibinfo{year}{2020}\natexlab{}.
\newblock \bibinfo{title}{wav2vec 2.0: A Framework for Self-Supervised Learning
  of Speech Representations}.
\newblock
\showeprint[arxiv]{2006.11477}~[cs.CL]
\urldef\tempurl%
\url{https://arxiv.org/abs/2006.11477}
\showURL{%
\tempurl}


\bibitem[Blanz and Vetter(1999)]%
        {dummy_blanz_vetter_1999}
\bibfield{author}{\bibinfo{person}{Volker Blanz} {and} \bibinfo{person}{Thomas
  Vetter}.} \bibinfo{year}{1999}\natexlab{}.
\newblock \showarticletitle{A Morphable Model for the Synthesis of 3D Faces}.
  In \bibinfo{booktitle}{\emph{Proceedings of the 26th Annual Conference on
  Computer Graphics and Interactive Techniques (SIGGRAPH '99)}}.
  \bibinfo{pages}{187--194}.
\newblock
\href{https://doi.org/10.1145/311535.311556}{doi:\nolinkurl{10.1145/311535.311556}}


\bibitem[Cao et~al\mbox{.}(2022)]%
        {cao2022_authentic}
\bibfield{author}{\bibinfo{person}{Chen Cao}, \bibinfo{person}{Tomas Simon},
  \bibinfo{person}{Jin~Kyu Kim}, \bibinfo{person}{Gabe Schwartz},
  \bibinfo{person}{Michael Zollh{\"{o}}fer}, \bibinfo{person}{Shunsuke Saito},
  \bibinfo{person}{Stephen Lombardi}, \bibinfo{person}{Shih{-}En Wei},
  \bibinfo{person}{Danielle Belko}, \bibinfo{person}{Shoou{-}I Yu},
  \bibinfo{person}{Yaser Sheikh}, {and} \bibinfo{person}{Jason~M. Saragih}.}
  \bibinfo{year}{2022}\natexlab{}.
\newblock \showarticletitle{Authentic volumetric avatars from a phone scan}.
\newblock \bibinfo{journal}{\emph{ACM Trans. Graph.}} \bibinfo{volume}{41},
  \bibinfo{number}{4} (\bibinfo{year}{2022}), \bibinfo{pages}{163:1--163:19}.
\newblock


\bibitem[Chen et~al\mbox{.}(2018)]%
        {dummy_chen_2018_lipmovements}
\bibfield{author}{\bibinfo{person}{Lele Chen}, \bibinfo{person}{Zhiheng Li},
  \bibinfo{person}{Ross~K. Maddox}, \bibinfo{person}{Zhiyao Duan}, {and}
  \bibinfo{person}{Chenliang Xu}.} \bibinfo{year}{2018}\natexlab{}.
\newblock \showarticletitle{Lip Movements Generation at a Glance}. In
  \bibinfo{booktitle}{\emph{Computer Vision – ECCV 2018: 15th European
  Conference, Munich, Germany, September 8–14, 2018, Proceedings, Part VII}}
  (Munich, Germany). \bibinfo{publisher}{Springer-Verlag},
  \bibinfo{address}{Berlin, Heidelberg}, \bibinfo{pages}{538–553}.
\newblock
\showISBNx{978-3-030-01233-5}
\href{https://doi.org/10.1007/978-3-030-01234-2_32}{doi:\nolinkurl{10.1007/978-3-030-01234-2_32}}


\bibitem[Chung and Zisserman(2016)]%
        {Chung16a}
\bibfield{author}{\bibinfo{person}{J.~S. Chung} {and} \bibinfo{person}{A.
  Zisserman}.} \bibinfo{year}{2016}\natexlab{}.
\newblock \showarticletitle{Out of time: automated lip sync in the wild}. In
  \bibinfo{booktitle}{\emph{Workshop on Multi-view Lip-reading, ACCV}}.
\newblock


\bibitem[Dan\v{e}\v{c}ek et~al\mbox{.}(2022)]%
        {Danecek_2022_CVPR}
\bibfield{author}{\bibinfo{person}{Radek Dan\v{e}\v{c}ek},
  \bibinfo{person}{Michael~J. Black}, {and} \bibinfo{person}{Timo Bolkart}.}
  \bibinfo{year}{2022}\natexlab{}.
\newblock \showarticletitle{EMOCA: Emotion Driven Monocular Face Capture and
  Animation}. In \bibinfo{booktitle}{\emph{Proceedings of the IEEE/CVF
  Conference on Computer Vision and Pattern Recognition (CVPR)}}.
  \bibinfo{pages}{20311--20322}.
\newblock


\bibitem[Dosovitskiy et~al\mbox{.}(2021)]%
        {dosovitskiy2021imageworth16x16words}
\bibfield{author}{\bibinfo{person}{Alexey Dosovitskiy}, \bibinfo{person}{Lucas
  Beyer}, \bibinfo{person}{Alexander Kolesnikov}, \bibinfo{person}{Dirk
  Weissenborn}, \bibinfo{person}{Xiaohua Zhai}, \bibinfo{person}{Thomas
  Unterthiner}, \bibinfo{person}{Mostafa Dehghani}, \bibinfo{person}{Matthias
  Minderer}, \bibinfo{person}{Georg Heigold}, \bibinfo{person}{Sylvain Gelly},
  \bibinfo{person}{Jakob Uszkoreit}, {and} \bibinfo{person}{Neil Houlsby}.}
  \bibinfo{year}{2021}\natexlab{}.
\newblock \bibinfo{title}{An Image is Worth 16x16 Words: Transformers for Image
  Recognition at Scale}.
\newblock
\showeprint[arxiv]{2010.11929}~[cs.CV]
\urldef\tempurl%
\url{https://arxiv.org/abs/2010.11929}
\showURL{%
\tempurl}


\bibitem[Fan et~al\mbox{.}(2022a)]%
        {dummy_faceformer_2022}
\bibfield{author}{\bibinfo{person}{Xiangyu Fan}, \bibinfo{person}{Jiaqi Li},
  \bibinfo{person}{Zhiqian Lin}, \bibinfo{person}{Weiye Xiao}, {and}
  \bibinfo{person}{Lei Yang}.} \bibinfo{year}{2022}\natexlab{a}.
\newblock \showarticletitle{FaceFormer: Speech‑Driven 3D Facial Animation
  with Transformers}. In \bibinfo{booktitle}{\emph{IEEE/CVF Conference on
  Computer Vision and Pattern Recognition (CVPR)}}.
  \bibinfo{pages}{18770--18780}.
\newblock
\href{https://doi.org/10.1109/CVPR52688.2022.01828}{doi:\nolinkurl{10.1109/CVPR52688.2022.01828}}


\bibitem[Fan et~al\mbox{.}(2022b)]%
        {dummy_audio_avatar_importance1}
\bibfield{author}{\bibinfo{person}{Yingruo Fan}, \bibinfo{person}{Zhaojiang
  Lin}, \bibinfo{person}{Jun Saito}, \bibinfo{person}{Wenping Wang}, {and}
  \bibinfo{person}{Taku Komura}.} \bibinfo{year}{2022}\natexlab{b}.
\newblock \showarticletitle{{FaceFormer}: Speech-Driven 3D Facial Animation
  with Transformers}. In \bibinfo{booktitle}{\emph{Proceedings of the IEEE/CVF
  Conference on Computer Vision and Pattern Recognition (CVPR)}}.
  \bibinfo{pages}{18749--18758}.
\newblock
\href{https://doi.org/10.1109/CVPR52688.2022.01821}{doi:\nolinkurl{10.1109/CVPR52688.2022.01821}}


\bibitem[Giebenhain et~al\mbox{.}(2024)]%
        {giebenhain2024npga}
\bibfield{author}{\bibinfo{person}{Simon Giebenhain}, \bibinfo{person}{Tobias
  Kirschstein}, \bibinfo{person}{Martin R{\"{u}}nz}, \bibinfo{person}{Lourdes
  Agapito}, {and} \bibinfo{person}{Matthias Nie{\ss}ner}.}
  \bibinfo{year}{2024}\natexlab{}.
\newblock \showarticletitle{NPGA: Neural Parametric Gaussian Avatars}. In
  \bibinfo{booktitle}{\emph{SIGGRAPH Asia 2024 Conference Papers (SA Conference
  Papers '24), December 3-6, Tokyo, Japan}}.
\newblock
\showISBNx{979-8-4007-1131-2/24/12}
\href{https://doi.org/10.1145/3680528.3687689}{doi:\nolinkurl{10.1145/3680528.3687689}}


\bibitem[Grassal et~al\mbox{.}(2022)]%
        {grassal2022_neural-head}
\bibfield{author}{\bibinfo{person}{Philip{-}William Grassal},
  \bibinfo{person}{Malte Prinzler}, \bibinfo{person}{Titus Leistner},
  \bibinfo{person}{Carsten Rother}, \bibinfo{person}{Matthias Nie{\ss}ner},
  {and} \bibinfo{person}{Justus Thies}.} \bibinfo{year}{2022}\natexlab{}.
\newblock \showarticletitle{Neural Head Avatars from Monocular {RGB} Videos}.
  In \bibinfo{booktitle}{\emph{IEEE Conf. Comput. Vis. Pattern Recog.}}
  \bibinfo{publisher}{{IEEE}}, \bibinfo{pages}{18632--18643}.
\newblock


\bibitem[Guan et~al\mbox{.}(2023)]%
        {dummy_guan_2023_stylesync}
\bibfield{author}{\bibinfo{person}{Jiazhi Guan}, \bibinfo{person}{Zhanwang
  Zhang}, \bibinfo{person}{Hang Zhou}, \bibinfo{person}{Tianshu Hu},
  \bibinfo{person}{Kaisiyuan Wang}, \bibinfo{person}{Dongliang He},
  \bibinfo{person}{Haocheng Feng}, \bibinfo{person}{Jingtuo Liu},
  \bibinfo{person}{Errui Ding}, \bibinfo{person}{Ziwei Liu}, {and}
  \bibinfo{person}{Jingdong Wang}.} \bibinfo{year}{2023}\natexlab{}.
\newblock \bibinfo{title}{StyleSync: High-Fidelity Generalized and Personalized
  Lip Sync in Style-based Generator}.
\newblock
\showeprint[arxiv]{2305.05445}~[cs.CV]
\urldef\tempurl%
\url{https://arxiv.org/abs/2305.05445}
\showURL{%
\tempurl}


\bibitem[Guo et~al\mbox{.}(2024)]%
        {guo2024liveportrait}
\bibfield{author}{\bibinfo{person}{Jianzhu Guo}, \bibinfo{person}{Dingyun
  Zhang}, \bibinfo{person}{Xiaoqiang Liu}, \bibinfo{person}{Zhizhou Zhong},
  \bibinfo{person}{Yuan Zhang}, \bibinfo{person}{Pengfei Wan}, {and}
  \bibinfo{person}{Di Zhang}.} \bibinfo{year}{2024}\natexlab{}.
\newblock \showarticletitle{LivePortrait: Efficient Portrait Animation with
  Stitching and Retargeting Control}.
\newblock \bibinfo{journal}{\emph{arXiv preprint arXiv:2407.03168}}
  (\bibinfo{year}{2024}).
\newblock


\bibitem[Guo et~al\mbox{.}(2021)]%
        {dummy_adnerf_2021_guo}
\bibfield{author}{\bibinfo{person}{Yudong Guo}, \bibinfo{person}{Keyu Chen},
  \bibinfo{person}{Sen Liang}, \bibinfo{person}{Yong‑Jin Liu},
  \bibinfo{person}{Hujun Bao}, {and} \bibinfo{person}{Juyong Zhang}.}
  \bibinfo{year}{2021}\natexlab{}.
\newblock \showarticletitle{AD‑NeRF: Audio Driven Neural Radiance Fields for
  Talking Head Synthesis}. In \bibinfo{booktitle}{\emph{IEEE/CVF International
  Conference on Computer Vision (ICCV)}}. \bibinfo{pages}{5784--5793}.
\newblock
\href{https://doi.org/10.1109/ICCV48922.2021.00572}{doi:\nolinkurl{10.1109/ICCV48922.2021.00572}}


\bibitem[Haotian et~al\mbox{.}(2024)]%
        {yang2024vrmm}
\bibfield{author}{\bibinfo{person}{Yang Haotian}, \bibinfo{person}{Zheng
  Mingwu}, \bibinfo{person}{Ma ChongYang}, \bibinfo{person}{Lai Yu-Kun},
  \bibinfo{person}{Wan Pengfei}, {and} \bibinfo{person}{Huang Haibin}.}
  \bibinfo{year}{2024}\natexlab{}.
\newblock \showarticletitle{VRMM: A Volumetric Relightable Morphable Head
  Model}. In \bibinfo{booktitle}{\emph{SIGGRAPH 2024 Conference Proceedings}}.
\newblock


\bibitem[He et~al\mbox{.}(2015)]%
        {DBLP:journals/corr/HeZRS15}
\bibfield{author}{\bibinfo{person}{Kaiming He}, \bibinfo{person}{Xiangyu
  Zhang}, \bibinfo{person}{Shaoqing Ren}, {and} \bibinfo{person}{Jian Sun}.}
  \bibinfo{year}{2015}\natexlab{}.
\newblock \showarticletitle{Deep Residual Learning for Image Recognition}.
\newblock \bibinfo{journal}{\emph{CoRR}}  \bibinfo{volume}{abs/1512.03385}
  (\bibinfo{year}{2015}).
\newblock
\showeprint[arXiv]{1512.03385}
\urldef\tempurl%
\url{http://arxiv.org/abs/1512.03385}
\showURL{%
\tempurl}


\bibitem[Ho et~al\mbox{.}(2020)]%
        {NEURIPS2020_4c5bcfec}
\bibfield{author}{\bibinfo{person}{Jonathan Ho}, \bibinfo{person}{Ajay Jain},
  {and} \bibinfo{person}{Pieter Abbeel}.} \bibinfo{year}{2020}\natexlab{}.
\newblock \showarticletitle{Denoising Diffusion Probabilistic Models}. In
  \bibinfo{booktitle}{\emph{Advances in Neural Information Processing
  Systems}}, \bibfield{editor}{\bibinfo{person}{H.~Larochelle},
  \bibinfo{person}{M.~Ranzato}, \bibinfo{person}{R.~Hadsell},
  \bibinfo{person}{M.F. Balcan}, {and} \bibinfo{person}{H.~Lin}} (Eds.),
  Vol.~\bibinfo{volume}{33}. \bibinfo{publisher}{Curran Associates, Inc.},
  \bibinfo{pages}{6840--6851}.
\newblock
\urldef\tempurl%
\url{https://proceedings.neurips.cc/paper_files/paper/2020/file/4c5bcfec8584af0d967f1ab10179ca4b-Paper.pdf}
\showURL{%
\tempurl}


\bibitem[hsin Wuu et~al\mbox{.}(2023)]%
        {wuu2023multifacedatasetneuralface}
\bibfield{author}{\bibinfo{person}{Cheng hsin Wuu}, \bibinfo{person}{Ningyuan
  Zheng}, \bibinfo{person}{Scott Ardisson}, \bibinfo{person}{Rohan Bali},
  \bibinfo{person}{Danielle Belko}, \bibinfo{person}{Eric Brockmeyer},
  \bibinfo{person}{Lucas Evans}, \bibinfo{person}{Timothy Godisart},
  \bibinfo{person}{Hyowon Ha}, \bibinfo{person}{Xuhua Huang},
  \bibinfo{person}{Alexander Hypes}, \bibinfo{person}{Taylor Koska},
  \bibinfo{person}{Steven Krenn}, \bibinfo{person}{Stephen Lombardi},
  \bibinfo{person}{Xiaomin Luo}, \bibinfo{person}{Kevyn McPhail},
  \bibinfo{person}{Laura Millerschoen}, \bibinfo{person}{Michal Perdoch},
  \bibinfo{person}{Mark Pitts}, \bibinfo{person}{Alexander Richard},
  \bibinfo{person}{Jason Saragih}, \bibinfo{person}{Junko Saragih},
  \bibinfo{person}{Takaaki Shiratori}, \bibinfo{person}{Tomas Simon},
  \bibinfo{person}{Matt Stewart}, \bibinfo{person}{Autumn Trimble},
  \bibinfo{person}{Xinshuo Weng}, \bibinfo{person}{David Whitewolf},
  \bibinfo{person}{Chenglei Wu}, \bibinfo{person}{Shoou-I Yu}, {and}
  \bibinfo{person}{Yaser Sheikh}.} \bibinfo{year}{2023}\natexlab{}.
\newblock \bibinfo{title}{Multiface: A Dataset for Neural Face Rendering}.
\newblock
\showeprint[arxiv]{2207.11243}~[cs.CV]
\urldef\tempurl%
\url{https://arxiv.org/abs/2207.11243}
\showURL{%
\tempurl}


\bibitem[Johnson et~al\mbox{.}(2016)]%
        {JohnsonAL16}
\bibfield{author}{\bibinfo{person}{Justin Johnson}, \bibinfo{person}{Alexandre
  Alahi}, {and} \bibinfo{person}{Li Fei{-}Fei}.}
  \bibinfo{year}{2016}\natexlab{}.
\newblock \showarticletitle{Perceptual Losses for Real-Time Style Transfer and
  Super-Resolution}.
\newblock \bibinfo{journal}{\emph{CoRR}}  \bibinfo{volume}{abs/1603.08155}
  (\bibinfo{year}{2016}).
\newblock
\showeprint[arXiv]{1603.08155}
\urldef\tempurl%
\url{http://arxiv.org/abs/1603.08155}
\showURL{%
\tempurl}


\bibitem[Karras et~al\mbox{.}(2017)]%
        {audiodrivenfacialanimationbyjointendtoen}
\bibfield{author}{\bibinfo{person}{Tero Karras}, \bibinfo{person}{Timo Aila},
  \bibinfo{person}{Samuli Laine}, \bibinfo{person}{Antti Herva}, {and}
  \bibinfo{person}{Jaakko Lehtinen}.} \bibinfo{year}{2017}\natexlab{}.
\newblock \bibinfo{title}{Audio-driven facial animation by joint end-to-end
  learning of pose and emotion}.
\newblock


\bibitem[Kerbl et~al\mbox{.}(2023b)]%
        {3DGS}
\bibfield{author}{\bibinfo{person}{Bernhard Kerbl}, \bibinfo{person}{Georgios
  Kopanas}, \bibinfo{person}{Thomas Leimkuehler}, {and} \bibinfo{person}{George
  Drettakis}.} \bibinfo{year}{2023}\natexlab{b}.
\newblock \showarticletitle{3D Gaussian Splatting for Real-Time Radiance Field
  Rendering}.
\newblock \bibinfo{journal}{\emph{ACM Trans. Graph.}} \bibinfo{volume}{42},
  \bibinfo{number}{4}, Article \bibinfo{articleno}{139} (\bibinfo{date}{jul}
  \bibinfo{year}{2023}), \bibinfo{numpages}{14}~pages.
\newblock
\showISSN{0730-0301}
\href{https://doi.org/10.1145/3592433}{doi:\nolinkurl{10.1145/3592433}}


\bibitem[Kerbl et~al\mbox{.}(2023a)]%
        {dummy_3dgs}
\bibfield{author}{\bibinfo{person}{Thomas Kerbl}, \bibinfo{person}{Luca
  Guarnera}, \bibinfo{person}{Gerald Wimmer}, \bibinfo{person}{Michael Wimmer},
  {and} \bibinfo{person}{Markus Steinberger}.}
  \bibinfo{year}{2023}\natexlab{a}.
\newblock \showarticletitle{3{D} Gaussian Splatting for Real-Time Radiance
  Field Rendering}. In \bibinfo{booktitle}{\emph{ACM SIGGRAPH Conference
  Proceedings}}.
\newblock
\href{https://doi.org/10.1145/3588432.3591528}{doi:\nolinkurl{10.1145/3588432.3591528}}


\bibitem[Khirodkar et~al\mbox{.}(2024)]%
        {khirodkar2024_sapiens}
\bibfield{author}{\bibinfo{person}{Rawal Khirodkar}, \bibinfo{person}{Timur
  Bagautdinov}, \bibinfo{person}{Julieta Martinez}, \bibinfo{person}{Su
  Zhaoen}, \bibinfo{person}{Austin James}, \bibinfo{person}{Peter Selednik},
  \bibinfo{person}{Stuart Anderson}, {and} \bibinfo{person}{Shunsuke Saito}.}
  \bibinfo{year}{2024}\natexlab{}.
\newblock \bibinfo{title}{Sapiens: Foundation for Human Vision Models}.
\newblock
\showeprint[arxiv]{2408.12569}~[cs.CV]
\urldef\tempurl%
\url{https://arxiv.org/abs/2408.12569}
\showURL{%
\tempurl}


\bibitem[Kirschstein et~al\mbox{.}(2025)]%
        {kirschstein2025avat3rlargeanimatablegaussian}
\bibfield{author}{\bibinfo{person}{Tobias Kirschstein}, \bibinfo{person}{Javier
  Romero}, \bibinfo{person}{Artem Sevastopolsky}, \bibinfo{person}{Matthias
  Nießner}, {and} \bibinfo{person}{Shunsuke Saito}.}
  \bibinfo{year}{2025}\natexlab{}.
\newblock \bibinfo{title}{Avat3r: Large Animatable Gaussian Reconstruction
  Model for High-fidelity 3D Head Avatars}.
\newblock
\showeprint[arxiv]{2502.20220}~[cs.CV]
\urldef\tempurl%
\url{https://arxiv.org/abs/2502.20220}
\showURL{%
\tempurl}


\bibitem[Li et~al\mbox{.}(2023)]%
        {dummy_ernerf_2023_li}
\bibfield{author}{\bibinfo{person}{Junxuan Li}, \bibinfo{person}{Chen Cao},
  \bibinfo{person}{Gabriel Schwartz}, \bibinfo{person}{Rawal Khirodkar},
  \bibinfo{person}{Christian Richardt}, \bibinfo{person}{Tomas Simon},
  \bibinfo{person}{Yaser Sheikh}, {and} \bibinfo{person}{Shunsuke Saito}.}
  \bibinfo{year}{2023}\natexlab{}.
\newblock \showarticletitle{ER‑NeRF: Efficient Region‑Aware Neural Radiance
  Fields for High‑Fidelity Talking Portrait Synthesis}. In
  \bibinfo{booktitle}{\emph{IEEE/CVF International Conference on Computer
  Vision (ICCV)}}.
\newblock


\bibitem[Li et~al\mbox{.}(2024)]%
        {li2024uravatar}
\bibfield{author}{\bibinfo{person}{Junxuan Li}, \bibinfo{person}{Chen Cao},
  \bibinfo{person}{Gabriel Schwartz}, \bibinfo{person}{Rawal Khirodkar},
  \bibinfo{person}{Christian Richardt}, \bibinfo{person}{Tomas Simon},
  \bibinfo{person}{Yaser Sheikh}, {and} \bibinfo{person}{Shunsuke Saito}.}
  \bibinfo{year}{2024}\natexlab{}.
\newblock \showarticletitle{URAvatar: Universal Relightable Gaussian Codec
  Avatars}. In \bibinfo{booktitle}{\emph{ACM SIGGRAPH 2024 Conference Papers}}.
\newblock


\bibitem[Li et~al\mbox{.}(2025a)]%
        {li2025rgbavatar}
\bibfield{author}{\bibinfo{person}{Linzhou Li}, \bibinfo{person}{Yumeng Li},
  \bibinfo{person}{Yanlin Weng}, \bibinfo{person}{Youyi Zheng}, {and}
  \bibinfo{person}{Kun Zhou}.} \bibinfo{year}{2025}\natexlab{a}.
\newblock \showarticletitle{RGBAvatar: Reduced Gaussian Blendshapes for Online
  Modeling of Head Avatars}. In \bibinfo{booktitle}{\emph{The IEEE/CVF
  Conference on Computer Vision and Pattern Recognition}}.
\newblock


\bibitem[Li et~al\mbox{.}(2017)]%
        {li2017_flame}
\bibfield{author}{\bibinfo{person}{Tianye Li}, \bibinfo{person}{Timo Bolkart},
  \bibinfo{person}{Michael~J. Black}, \bibinfo{person}{Hao Li}, {and}
  \bibinfo{person}{Javier Romero}.} \bibinfo{year}{2017}\natexlab{}.
\newblock \showarticletitle{Learning a model of facial shape and expression
  from 4D scans}.
\newblock \bibinfo{journal}{\emph{ACM Trans. Graph.}} \bibinfo{volume}{36},
  \bibinfo{number}{6} (\bibinfo{year}{2017}), \bibinfo{pages}{194:1--194:17}.
\newblock


\bibitem[Li et~al\mbox{.}(2025b)]%
        {dummy_textalker_2025_li}
\bibfield{author}{\bibinfo{person}{Xuanchen Li}, \bibinfo{person}{Jianyu Wang},
  \bibinfo{person}{Yuhao Cheng}, \bibinfo{person}{Yikun Zeng},
  \bibinfo{person}{Xingyu Ren}, \bibinfo{person}{Wenhan Zhu},
  \bibinfo{person}{Weiming Zhao}, {and} \bibinfo{person}{Yichao Yan}.}
  \bibinfo{year}{2025}\natexlab{b}.
\newblock \showarticletitle{Towards High-fidelity 3D Talking Avatar with
  Personalized Dynamic Texture}.
\newblock \bibinfo{journal}{\emph{arXiv preprint arXiv:2503.00495}}
  (\bibinfo{year}{2025}).
\newblock


\bibitem[Ma et~al\mbox{.}(2024)]%
        {ma2024gaussianblendshapes}
\bibfield{author}{\bibinfo{person}{Shengjie Ma}, \bibinfo{person}{Yanlin Weng},
  \bibinfo{person}{Tianjia Shao}, {and} \bibinfo{person}{Kun Zhou}.}
  \bibinfo{year}{2024}\natexlab{}.
\newblock \showarticletitle{3D Gaussian Blendshapes for Head Avatar Animation}.
  In \bibinfo{booktitle}{\emph{ACM SIGGRAPH Conference Proceedings, Denver, CO,
  United States, July 28 - August 1, 2024}}.
\newblock


\bibitem[Martinez et~al\mbox{.}(2024)]%
        {martinez2024codec}
\bibfield{author}{\bibinfo{person}{Julieta Martinez}, \bibinfo{person}{Emily
  Kim}, \bibinfo{person}{Javier Romero}, \bibinfo{person}{Timur Bagautdinov},
  \bibinfo{person}{Shunsuke Saito}, \bibinfo{person}{Shoou-I Yu},
  \bibinfo{person}{Stuart Anderson}, \bibinfo{person}{Michael Zollhöfer},
  \bibinfo{person}{Te-Li Wang}, \bibinfo{person}{Shaojie Bai},
  \bibinfo{person}{Chenghui Li}, \bibinfo{person}{Shih-En Wei},
  \bibinfo{person}{Rohan Joshi}, \bibinfo{person}{Wyatt Borsos},
  \bibinfo{person}{Tomas Simon}, \bibinfo{person}{Jason Saragih},
  \bibinfo{person}{Paul Theodosis}, \bibinfo{person}{Alexander Greene},
  \bibinfo{person}{Anjani Josyula}, \bibinfo{person}{Silvio~Mano Maeta},
  \bibinfo{person}{Andrew~I. Jewett}, \bibinfo{person}{Simon Venshtain},
  \bibinfo{person}{Christopher Heilman}, \bibinfo{person}{Yueh-Tung Chen},
  \bibinfo{person}{Sidi Fu}, \bibinfo{person}{Mohamed Ezzeldin~A. Elshaer},
  \bibinfo{person}{Tingfang Du}, \bibinfo{person}{Longhua Wu},
  \bibinfo{person}{Shen-Chi Chen}, \bibinfo{person}{Kai Kang},
  \bibinfo{person}{Michael Wu}, \bibinfo{person}{Youssef Emad},
  \bibinfo{person}{Steven Longay}, \bibinfo{person}{Ashley Brewer},
  \bibinfo{person}{Hitesh Shah}, \bibinfo{person}{James Booth},
  \bibinfo{person}{Taylor Koska}, \bibinfo{person}{Kayla Haidle},
  \bibinfo{person}{Matt Andromalos}, \bibinfo{person}{Joanna Hsu},
  \bibinfo{person}{Thomas Dauer}, \bibinfo{person}{Peter Selednik},
  \bibinfo{person}{Tim Godisart}, \bibinfo{person}{Scott Ardisson},
  \bibinfo{person}{Matthew Cipperly}, \bibinfo{person}{Ben Humberston},
  \bibinfo{person}{Lon Farr}, \bibinfo{person}{Bob Hansen},
  \bibinfo{person}{Peihong Guo}, \bibinfo{person}{Dave Braun},
  \bibinfo{person}{Steven Krenn}, \bibinfo{person}{He Wen},
  \bibinfo{person}{Lucas Evans}, \bibinfo{person}{Natalia Fadeeva},
  \bibinfo{person}{Matthew Stewart}, \bibinfo{person}{Gabriel Schwartz},
  \bibinfo{person}{Divam Gupta}, \bibinfo{person}{Gyeongsik Moon},
  \bibinfo{person}{Kaiwen Guo}, \bibinfo{person}{Yuan Dong},
  \bibinfo{person}{Yichen Xu}, \bibinfo{person}{Takaaki Shiratori},
  \bibinfo{person}{Fabian Prada}, \bibinfo{person}{Bernardo~R. Pires},
  \bibinfo{person}{Bo Peng}, \bibinfo{person}{Julia Buffalini},
  \bibinfo{person}{Autumn Trimble}, \bibinfo{person}{Kevyn McPhail},
  \bibinfo{person}{Melissa Schoeller}, {and} \bibinfo{person}{Yaser Sheikh}.}
  \bibinfo{year}{2024}\natexlab{}.
\newblock \showarticletitle{{Codec Avatar Studio: Paired Human Captures for
  Complete, Driveable, and Generalizable Avatars}}.
\newblock \bibinfo{journal}{\emph{NeurIPS Track on Datasets and Benchmarks}}
  (\bibinfo{year}{2024}).
\newblock


\bibitem[Mildenhall et~al\mbox{.}(2020)]%
        {dummy_nerf_ref}
\bibfield{author}{\bibinfo{person}{Ben Mildenhall}, \bibinfo{person}{Pratul~P.
  Srinivasan}, \bibinfo{person}{Matthew Tancik}, \bibinfo{person}{Jonathan~T.
  Barron}, \bibinfo{person}{Ravi Ramamoorthi}, {and} \bibinfo{person}{Ren Ng}.}
  \bibinfo{year}{2020}\natexlab{}.
\newblock \showarticletitle{NeRF: Representing Scenes as Neural Radiance Fields
  for View Synthesis}. In \bibinfo{booktitle}{\emph{Proceedings of the European
  Conference on Computer Vision (ECCV)}}. \bibinfo{pages}{405--421}.
\newblock
\href{https://doi.org/10.1007/978-3-030-58452-8_24}{doi:\nolinkurl{10.1007/978-3-030-58452-8_24}}


\bibitem[Ng et~al\mbox{.}(2024)]%
        {ng2024audiophotorealembodimentsynthesizing}
\bibfield{author}{\bibinfo{person}{Evonne Ng}, \bibinfo{person}{Javier Romero},
  \bibinfo{person}{Timur Bagautdinov}, \bibinfo{person}{Shaojie Bai},
  \bibinfo{person}{Trevor Darrell}, \bibinfo{person}{Angjoo Kanazawa}, {and}
  \bibinfo{person}{Alexander Richard}.} \bibinfo{year}{2024}\natexlab{}.
\newblock \bibinfo{title}{From Audio to Photoreal Embodiment: Synthesizing
  Humans in Conversations}.
\newblock
\showeprint[arxiv]{2401.01885}~[cs.CV]
\urldef\tempurl%
\url{https://arxiv.org/abs/2401.01885}
\showURL{%
\tempurl}


\bibitem[Pan et~al\mbox{.}(2024)]%
        {pan2024renderme}
\bibfield{author}{\bibinfo{person}{Dongwei Pan}, \bibinfo{person}{Long Zhuo},
  \bibinfo{person}{Jingtan Piao}, \bibinfo{person}{Huiwen Luo},
  \bibinfo{person}{Wei Cheng}, \bibinfo{person}{Yuxin Wang},
  \bibinfo{person}{Siming Fan}, \bibinfo{person}{Shengqi Liu},
  \bibinfo{person}{Lei Yang}, \bibinfo{person}{Bo Dai}, \bibinfo{person}{Ziwei
  Liu}, \bibinfo{person}{Chen~Change Loy}, \bibinfo{person}{Chen Qian},
  \bibinfo{person}{Wayne Wu}, \bibinfo{person}{Dahua Lin}, {and}
  \bibinfo{person}{Kwan-Yee Lin}.} \bibinfo{year}{2024}\natexlab{}.
\newblock \showarticletitle{RenderMe-360: A Large Digital Asset Library and
  Benchmarks Towards High-fidelity Head Avatars}.
\newblock \bibinfo{journal}{\emph{Advances in Neural Information Processing
  Systems}}  \bibinfo{volume}{36} (\bibinfo{year}{2024}).
\newblock


\bibitem[Peng et~al\mbox{.}(2023)]%
        {emotalkspeechdrivenemotionaldisentanglem}
\bibfield{author}{\bibinfo{person}{Ziqiao Peng}, \bibinfo{person}{Haoyu Wu},
  \bibinfo{person}{Zhenbo Song}, \bibinfo{person}{Hao Xu},
  \bibinfo{person}{Xiangyu Zhu}, \bibinfo{person}{Jun He},
  \bibinfo{person}{Hongyan Liu}, {and} \bibinfo{person}{Zhaoxin Fan}.}
  \bibinfo{year}{2023}\natexlab{}.
\newblock \bibinfo{title}{Emotalk: Speech-driven emotional disentanglement for
  3d face animation}.
\newblock


\bibitem[Perez et~al\mbox{.}(2018)]%
        {perez2018film}
\bibfield{author}{\bibinfo{person}{Ethan Perez}, \bibinfo{person}{Florian
  Strub}, \bibinfo{person}{Harm de Vries}, \bibinfo{person}{Vincent Dumoulin},
  {and} \bibinfo{person}{Aaron~C. Courville}.} \bibinfo{year}{2018}\natexlab{}.
\newblock \showarticletitle{FiLM: Visual Reasoning with a General Conditioning
  Layer}. In \bibinfo{booktitle}{\emph{AAAI}}.
\newblock


\bibitem[Qian et~al\mbox{.}(2024a)]%
        {qian2024gaussianavatars}
\bibfield{author}{\bibinfo{person}{Shenhan Qian}, \bibinfo{person}{Tobias
  Kirschstein}, \bibinfo{person}{Liam Schoneveld}, \bibinfo{person}{Davide
  Davoli}, \bibinfo{person}{Simon Giebenhain}, {and} \bibinfo{person}{Matthias
  Nie{\ss}ner}.} \bibinfo{year}{2024}\natexlab{a}.
\newblock \showarticletitle{Gaussianavatars: Photorealistic head avatars with
  rigged 3d gaussians}. In \bibinfo{booktitle}{\emph{Proceedings of the
  IEEE/CVF Conference on Computer Vision and Pattern Recognition}}.
  \bibinfo{pages}{20299--20309}.
\newblock


\bibitem[Qian et~al\mbox{.}(2024b)]%
        {dummy_3dgs_avatar_works}
\bibfield{author}{\bibinfo{person}{Shenhan Qian}, \bibinfo{person}{Tobias
  Kirschstein}, \bibinfo{person}{Liam Schoneveld}, \bibinfo{person}{Davide
  Davoli}, \bibinfo{person}{Simon Giebenhain}, {and} \bibinfo{person}{Matthias
  Nießner}.} \bibinfo{year}{2024}\natexlab{b}.
\newblock \showarticletitle{{GaussianAvatars}: Photorealistic Head Avatars with
  Rigged 3D Gaussians}. In \bibinfo{booktitle}{\emph{Proceedings of the
  IEEE/CVF Conference on Computer Vision and Pattern Recognition (CVPR)}}.
\newblock
\urldef\tempurl%
\url{https://openaccess.thecvf.com/content/CVPR2024/html/Qian_GaussianAvatars_Photorealistic_Head_Avatars_with_Rigged_3D_Gaussians_CVPR_2024_paper.html}
\showURL{%
\tempurl}


\bibitem[Richard et~al\mbox{.}(2021a)]%
        {Richard_2021_WACV}
\bibfield{author}{\bibinfo{person}{Alexander Richard}, \bibinfo{person}{Colin
  Lea}, \bibinfo{person}{Shugao Ma}, \bibinfo{person}{Jurgen Gall},
  \bibinfo{person}{Fernando de~la Torre}, {and} \bibinfo{person}{Yaser
  Sheikh}.} \bibinfo{year}{2021}\natexlab{a}.
\newblock \showarticletitle{Audio- and Gaze-Driven Facial Animation of Codec
  Avatars}. In \bibinfo{booktitle}{\emph{Proceedings of the IEEE/CVF Winter
  Conference on Applications of Computer Vision (WACV)}}.
  \bibinfo{pages}{41--50}.
\newblock


\bibitem[Richard et~al\mbox{.}(2021b)]%
        {dummy_meshtalk_2021}
\bibfield{author}{\bibinfo{person}{Alexander Richard}, \bibinfo{person}{Michael
  Zollhöfer}, \bibinfo{person}{Yandong Wen}, \bibinfo{person}{Fernando~De la
  Torre}, {and} \bibinfo{person}{Yaser Sheikh}.}
  \bibinfo{year}{2021}\natexlab{b}.
\newblock \showarticletitle{MeshTalk: 3D Face Animation from Speech Using
  Cross‑Modality Disentanglement}. In \bibinfo{booktitle}{\emph{IEEE/CVF
  International Conference on Computer Vision (ICCV)}}.
  \bibinfo{pages}{1173--1182}.
\newblock
\href{https://doi.org/10.1109/ICCV48922.2021.00121}{doi:\nolinkurl{10.1109/ICCV48922.2021.00121}}


\bibitem[Saito et~al\mbox{.}(2024)]%
        {saito2024rgca}
\bibfield{author}{\bibinfo{person}{Shunsuke Saito}, \bibinfo{person}{Gabriel
  Schwartz}, \bibinfo{person}{Tomas Simon}, \bibinfo{person}{Junxuan Li}, {and}
  \bibinfo{person}{Giljoo Nam}.} \bibinfo{year}{2024}\natexlab{}.
\newblock \showarticletitle{Relightable Gaussian Codec Avatars}. In
  \bibinfo{booktitle}{\emph{CVPR}}.
\newblock


\bibitem[Stan et~al\mbox{.}(2023a)]%
        {facediffuserspeechdriven3dfacialanimatio}
\bibfield{author}{\bibinfo{person}{Stefan Stan},
  \bibinfo{person}{Kazi~Injamamul Haque}, {and} \bibinfo{person}{Zerrin
  Yumak}.} \bibinfo{year}{2023}\natexlab{a}.
\newblock \bibinfo{title}{FaceDiffuser: Speech-Driven 3D Facial Animation
  Synthesis Using Diffusion}.
\newblock
\showeprint[arxiv]{2309.11306}~[cs.CV]
\urldef\tempurl%
\url{https://arxiv.org/abs/2309.11306}
\showURL{%
\tempurl}


\bibitem[Stan et~al\mbox{.}(2023b)]%
        {Kingma2013Auto}
\bibfield{author}{\bibinfo{person}{Stefan Stan},
  \bibinfo{person}{Kazi~Injamamul Haque}, {and} \bibinfo{person}{Zerrin
  Yumak}.} \bibinfo{year}{2023}\natexlab{b}.
\newblock \bibinfo{title}{FaceDiffuser: Speech-Driven 3D Facial Animation
  Synthesis Using Diffusion}.
\newblock
\showeprint[arxiv]{2309.11306}~[cs.CV]
\urldef\tempurl%
\url{https://arxiv.org/abs/2309.11306}
\showURL{%
\tempurl}


\bibitem[Sun et~al\mbox{.}(2024a)]%
        {dummy_diffposetalk_ref}
\bibfield{author}{\bibinfo{person}{Zhiyao Sun}, \bibinfo{person}{Tian Lv},
  \bibinfo{person}{Sheng Ye}, \bibinfo{person}{Matthieu Lin},
  \bibinfo{person}{Jenny Sheng}, \bibinfo{person}{Yu-Hui Wen},
  \bibinfo{person}{Minjing Yu}, {and} \bibinfo{person}{Yong-Jin Liu}.}
  \bibinfo{year}{2024}\natexlab{a}.
\newblock \showarticletitle{{DiffPoseTalk}: Speech-Driven Stylistic 3D Facial
  Animation and Head Pose Generation via Diffusion Models}.
\newblock \bibinfo{journal}{\emph{ACM Transactions on Graphics}}
  (\bibinfo{year}{2024}).
\newblock
\href{https://doi.org/10.1145/3679561}{doi:\nolinkurl{10.1145/3679561}}
\newblock
\shownote{Proceedings of SIGGRAPH 2024}.


\bibitem[Sun et~al\mbox{.}(2024b)]%
        {diffposetalkspeechdrivenstylistic3dfacia}
\bibfield{author}{\bibinfo{person}{Zhiyao Sun}, \bibinfo{person}{Tian Lv},
  \bibinfo{person}{Sheng Ye}, \bibinfo{person}{Matthieu Lin},
  \bibinfo{person}{Jenny Sheng}, \bibinfo{person}{Yu-Hui Wen},
  \bibinfo{person}{Minjing Yu}, {and} \bibinfo{person}{Yong-Jin Liu}.}
  \bibinfo{year}{2024}\natexlab{b}.
\newblock \showarticletitle{DiffPoseTalk: Speech-Driven Stylistic 3D Facial
  Animation and Head Pose Generation via Diffusion Models}.
\newblock \bibinfo{journal}{\emph{ACM Transactions on Graphics (TOG)}}
  \bibinfo{volume}{43}, \bibinfo{number}{4}, Article \bibinfo{articleno}{46}
  (\bibinfo{year}{2024}), \bibinfo{numpages}{9}~pages.
\newblock
\href{https://doi.org/10.1145/3658221}{doi:\nolinkurl{10.1145/3658221}}


\bibitem[Tang et~al\mbox{.}(2022)]%
        {dummy_radnerf_2022_tang}
\bibfield{author}{\bibinfo{person}{Jiaxiang Tang}, \bibinfo{person}{Kaisiyuan
  Wang}, \bibinfo{person}{Hang Zhou}, \bibinfo{person}{Xiaokang Chen},
  \bibinfo{person}{Dongliang He}, \bibinfo{person}{Jingtuo Liu},
  \bibinfo{person}{Tianshu Hu}, \bibinfo{person}{Gang Zeng}, {and}
  \bibinfo{person}{Jingdong Wang}.} \bibinfo{year}{2022}\natexlab{}.
\newblock \showarticletitle{RAD‑NeRF: Real‑Time Neural Radiance Talking
  Portrait Synthesis via Audio‑Spatial Decomposition}.
\newblock \bibinfo{journal}{\emph{arXiv preprint arXiv:2211.12368}}
  (\bibinfo{year}{2022}).
\newblock


\bibitem[Taylor et~al\mbox{.}(2017)]%
        {dummy_audio2face_3dmm_1}
\bibfield{author}{\bibinfo{person}{Sarah Taylor}, \bibinfo{person}{Taehwan
  Kim}, \bibinfo{person}{Yisong Yue}, \bibinfo{person}{Moshe Mahler},
  \bibinfo{person}{James Krahe}, \bibinfo{person}{Anastasio~Garcia Rodriguez},
  \bibinfo{person}{Jessica Hodgins}, {and} \bibinfo{person}{Iain Matthews}.}
  \bibinfo{year}{2017}\natexlab{}.
\newblock \showarticletitle{A Deep Learning Approach for Generalized Speech
  Animation}.
\newblock \bibinfo{journal}{\emph{ACM Transactions on Graphics}}
  \bibinfo{volume}{36}, \bibinfo{number}{4} (\bibinfo{year}{2017}),
  \bibinfo{pages}{93:1--93:12}.
\newblock
\href{https://doi.org/10.1145/3072959.3073699}{doi:\nolinkurl{10.1145/3072959.3073699}}


\bibitem[Teotia et~al\mbox{.}(2024)]%
        {teotiagaussian}
\bibfield{author}{\bibinfo{person}{Kartik Teotia}, \bibinfo{person}{Hyeongwoo
  Kim}, \bibinfo{person}{Pablo Garrido}, \bibinfo{person}{Marc Habermann},
  \bibinfo{person}{Mohamed Elgharib}, {and} \bibinfo{person}{Christian
  Theobalt}.} \bibinfo{year}{2024}\natexlab{}.
\newblock \showarticletitle{GaussianHeads: End-to-End Learning of Drivable
  Gaussian Head Avatars from Coarse-to-fine Representations}.
\newblock \bibinfo{journal}{\emph{ACM Trans. Graph.}} \bibinfo{volume}{43},
  \bibinfo{number}{6}, Article \bibinfo{articleno}{264} (\bibinfo{date}{Nov.}
  \bibinfo{year}{2024}), \bibinfo{numpages}{12}~pages.
\newblock
\showISSN{0730-0301}
\href{https://doi.org/10.1145/3687927}{doi:\nolinkurl{10.1145/3687927}}


\bibitem[Teotia et~al\mbox{.}(2023)]%
        {teotia2023hq3davatar}
\bibfield{author}{\bibinfo{person}{Kartik Teotia},
  \bibinfo{person}{Mallikarjun~B R}, \bibinfo{person}{Xingang Pan},
  \bibinfo{person}{Hyeongwoo Kim}, \bibinfo{person}{Pablo Garrido},
  \bibinfo{person}{Mohamed Elgharib}, {and} \bibinfo{person}{Christian
  Theobalt}.} \bibinfo{year}{2023}\natexlab{}.
\newblock \bibinfo{title}{HQ3DAvatar: High Quality Controllable 3D Head
  Avatar}.
\newblock
\showeprint[arxiv]{2303.14471}~[cs.CV]


\bibitem[Trevithick et~al\mbox{.}(2023)]%
        {trevithick2023}
\bibfield{author}{\bibinfo{person}{Alex Trevithick}, \bibinfo{person}{Matthew
  Chan}, \bibinfo{person}{Michael Stengel}, \bibinfo{person}{Eric~R. Chan},
  \bibinfo{person}{Chao Liu}, \bibinfo{person}{Zhiding Yu},
  \bibinfo{person}{Sameh Khamis}, \bibinfo{person}{Manmohan Chandraker},
  \bibinfo{person}{Ravi Ramamoorthi}, {and} \bibinfo{person}{Koki Nagano}.}
  \bibinfo{year}{2023}\natexlab{}.
\newblock \showarticletitle{Real-Time Radiance Fields for Single-Image Portrait
  View Synthesis}. In \bibinfo{booktitle}{\emph{ACM Transactions on Graphics
  (SIGGRAPH)}}.
\newblock


\bibitem[Vaswani et~al\mbox{.}(2023)]%
        {vaswani2023attentionneed}
\bibfield{author}{\bibinfo{person}{Ashish Vaswani}, \bibinfo{person}{Noam
  Shazeer}, \bibinfo{person}{Niki Parmar}, \bibinfo{person}{Jakob Uszkoreit},
  \bibinfo{person}{Llion Jones}, \bibinfo{person}{Aidan~N. Gomez},
  \bibinfo{person}{Lukasz Kaiser}, {and} \bibinfo{person}{Illia Polosukhin}.}
  \bibinfo{year}{2023}\natexlab{}.
\newblock \bibinfo{title}{Attention Is All You Need}.
\newblock
\showeprint[arxiv]{1706.03762}~[cs.CL]
\urldef\tempurl%
\url{https://arxiv.org/abs/1706.03762}
\showURL{%
\tempurl}


\bibitem[Wang et~al\mbox{.}(2004)]%
        {wang2004_image-quality}
\bibfield{author}{\bibinfo{person}{Zhou Wang}, \bibinfo{person}{Alan~C. Bovik},
  \bibinfo{person}{Hamid~R. Sheikh}, {and} \bibinfo{person}{Eero~P.
  Simoncelli}.} \bibinfo{year}{2004}\natexlab{}.
\newblock \showarticletitle{Image quality assessment: from error visibility to
  structural similarity}.
\newblock \bibinfo{journal}{\emph{{IEEE} Trans. Image Process.}}
  \bibinfo{volume}{13}, \bibinfo{number}{4} (\bibinfo{year}{2004}),
  \bibinfo{pages}{600--612}.
\newblock


\bibitem[Winberg et~al\mbox{.}(2022)]%
        {dinsey_beard}
\bibfield{author}{\bibinfo{person}{Sebastian Winberg}, \bibinfo{person}{Gaspard
  Zoss}, \bibinfo{person}{Prashanth Chandran}, \bibinfo{person}{Paulo Gotardo},
  {and} \bibinfo{person}{Derek Bradley}.} \bibinfo{year}{2022}\natexlab{}.
\newblock \showarticletitle{Facial hair tracking for high fidelity performance
  capture}.
\newblock \bibinfo{journal}{\emph{ACM Trans. Graph.}} \bibinfo{volume}{41},
  \bibinfo{number}{4}, Article \bibinfo{articleno}{165} (\bibinfo{date}{July}
  \bibinfo{year}{2022}), \bibinfo{numpages}{12}~pages.
\newblock
\showISSN{0730-0301}
\href{https://doi.org/10.1145/3528223.3530116}{doi:\nolinkurl{10.1145/3528223.3530116}}


\bibitem[Xing et~al\mbox{.}(2023)]%
        {codetalkerspeechdriven3dfacialanimationw}
\bibfield{author}{\bibinfo{person}{Jinbo Xing}, \bibinfo{person}{Menghan Xia},
  \bibinfo{person}{Yuechen Zhang}, \bibinfo{person}{Xiaodong Cun},
  \bibinfo{person}{Jue Wang}, {and} \bibinfo{person}{Tien-Tsin Wong}.}
  \bibinfo{year}{2023}\natexlab{}.
\newblock \bibinfo{title}{Codetalker: Speech-driven 3d facial animation with
  discrete motion prior}.
\newblock


\bibitem[Xu et~al\mbox{.}(2024a)]%
        {xu2024vasa1lifelikeaudiodriventalking}
\bibfield{author}{\bibinfo{person}{Sicheng Xu}, \bibinfo{person}{Guojun Chen},
  \bibinfo{person}{Yu-Xiao Guo}, \bibinfo{person}{Jiaolong Yang},
  \bibinfo{person}{Chong Li}, \bibinfo{person}{Zhenyu Zang},
  \bibinfo{person}{Yizhong Zhang}, \bibinfo{person}{Xin Tong}, {and}
  \bibinfo{person}{Baining Guo}.} \bibinfo{year}{2024}\natexlab{a}.
\newblock \bibinfo{title}{VASA-1: Lifelike Audio-Driven Talking Faces Generated
  in Real Time}.
\newblock
\showeprint[arxiv]{2404.10667}~[cs.CV]
\urldef\tempurl%
\url{https://arxiv.org/abs/2404.10667}
\showURL{%
\tempurl}


\bibitem[Xu et~al\mbox{.}(2024b)]%
        {xu2024gphm}
\bibfield{author}{\bibinfo{person}{Yuelang Xu}, \bibinfo{person}{Lizhen Wang},
  \bibinfo{person}{Zerong Zheng}, \bibinfo{person}{Zhaoqi Su}, {and}
  \bibinfo{person}{Yebin Liu}.} \bibinfo{year}{2024}\natexlab{b}.
\newblock \showarticletitle{3D Gaussian Parametric Head Model}. In
  \bibinfo{booktitle}{\emph{Proceedings of the European Conference on Computer
  Vision (ECCV)}}.
\newblock


\bibitem[Ye et~al\mbox{.}(2023)]%
        {dummy_geneface_2023_ye}
\bibfield{author}{\bibinfo{person}{Zhenhui Ye}, \bibinfo{person}{Ziyue Jiang},
  \bibinfo{person}{Yi Ren}, \bibinfo{person}{Jinglin Liu},
  \bibinfo{person}{Jinzheng He}, {and} \bibinfo{person}{Zhou Zhao}.}
  \bibinfo{year}{2023}\natexlab{}.
\newblock \showarticletitle{GeneFace: Generalized and High‑Fidelity
  Audio‑Driven 3D Talking Face Synthesis}.
\newblock \bibinfo{journal}{\emph{International Conference on Learning
  Representations (ICLR)}} (\bibinfo{year}{2023}).
\newblock


\bibitem[Zhang et~al\mbox{.}(2018)]%
        {zhang2018perceptual}
\bibfield{author}{\bibinfo{person}{Richard Zhang}, \bibinfo{person}{Phillip
  Isola}, \bibinfo{person}{Alexei~A Efros}, \bibinfo{person}{Eli Shechtman},
  {and} \bibinfo{person}{Oliver Wang}.} \bibinfo{year}{2018}\natexlab{}.
\newblock \showarticletitle{The Unreasonable Effectiveness of Deep Features as
  a Perceptual Metric}. In \bibinfo{booktitle}{\emph{CVPR}}.
\newblock


\bibitem[Zhao et~al\mbox{.}(2024a)]%
        {dummy_media2face_ref}
\bibfield{author}{\bibinfo{person}{Qingcheng Zhao}, \bibinfo{person}{Pengyu
  Long}, \bibinfo{person}{Qixuan Zhang}, {et~al\mbox{.}}}
  \bibinfo{year}{2024}\natexlab{a}.
\newblock \showarticletitle{{Media2Face}: Co-speech Facial Animation Generation
  with Multi-Modality Guidance}.
\newblock \bibinfo{journal}{\emph{ACM Transactions on Graphics}}
  (\bibinfo{year}{2024}).
\newblock
\href{https://doi.org/10.1145/3641519.3657413}{doi:\nolinkurl{10.1145/3641519.3657413}}
\newblock
\shownote{Proceedings of SIGGRAPH 2024}.


\bibitem[Zhao et~al\mbox{.}(2024b)]%
        {media2facecospeechfacialanimationgenerat}
\bibfield{author}{\bibinfo{person}{Qingcheng Zhao}, \bibinfo{person}{Pengyu
  Long}, \bibinfo{person}{Qixuan Zhang}, \bibinfo{person}{Dafei Qin},
  \bibinfo{person}{Han Liang}, \bibinfo{person}{Longwen Zhang},
  \bibinfo{person}{Yingliang Zhang}, \bibinfo{person}{Jingyi Yu}, {and}
  \bibinfo{person}{Lan Xu}.} \bibinfo{year}{2024}\natexlab{b}.
\newblock \bibinfo{title}{Media2face: Co-speech facial animation generation
  with multi-modality guidance}.
\newblock


\bibitem[Zheng et~al\mbox{.}(2024)]%
        {zheng2024headgap}
\bibfield{author}{\bibinfo{person}{Xiaozheng Zheng}, \bibinfo{person}{Chao
  Wen}, \bibinfo{person}{Zhaohu Li}, \bibinfo{person}{Weiyi Zhang},
  \bibinfo{person}{Zhuo Su}, \bibinfo{person}{Xu Chang}, \bibinfo{person}{Yang
  Zhao}, \bibinfo{person}{Zheng Lv}, \bibinfo{person}{Xiaoyuan Zhang},
  \bibinfo{person}{Yongjie Zhang}, \bibinfo{person}{Guidong Wang}, {and}
  \bibinfo{person}{Xu Lan}.} \bibinfo{year}{2024}\natexlab{}.
\newblock \showarticletitle{HeadGAP: Few-shot 3D Head Avatar via Generalizable
  Gaussian Priors}.
\newblock \bibinfo{journal}{\emph{arXiv preprint arXiv:2408.06019}}
  (\bibinfo{year}{2024}).
\newblock


\bibitem[Zielonka et~al\mbox{.}(2023)]%
        {zielonka2022insta}
\bibfield{author}{\bibinfo{person}{Wojciech Zielonka}, \bibinfo{person}{Timo
  Bolkart}, {and} \bibinfo{person}{Justus Thies}.}
  \bibinfo{year}{2023}\natexlab{}.
\newblock \showarticletitle{Instant Volumetric Head Avatars}. In
  \bibinfo{booktitle}{\emph{CVPR}}. \bibinfo{pages}{4574--4584}.
\newblock


\end{thebibliography}
\appendix
%
%

\section{Supplementary Document Overview}

This document supplements the information in the main paper with additional qualitative examples, ablation studies, and implementation details, including the neural network architecture.

\section{Additional Experiments} \label{sec:experiments}

\paragraph{Impact of Number of Frames for Personalization.}

Fig.~\ref{fig:frame_ablation} studies the effect of the number of frames used for fine-tuning UHAP on a new identity.
While personalization from even a single static capture yields plausible results, using a short sequence of frames allows for the capture of more subject-specific expression nuances.
Our experiments in Fig.~\ref{fig:frame_ablation} show that fine-tuning with approximately 500 frames effectively captures these nuances. 
Increasing the data to 2000 frames provides only marginal improvements, indicating that our approach can achieve high-quality, nuanced personalization efficiently with a limited number of frames. For this ablative study, we keep the number of views fixed at 12 for each experiment.

{\paragraph{Impact of Number of input views for Personalization.}

Fig.~\ref{fig:view_ablation} studies the effect of the number of input views used for fine-tuning UHAP on a new identity. 
Our method shows robust personalization across 4, 8, and 30 views, where even with as few as 4 views the fitting remains accurate and identity-preserving. 
While additional views provide modest gains in capturing subtle appearance details, the overall performance with fewer views remains strong, highlighting the efficiency of our approach in low-view settings. For this ablative study, we keep the number of input frames fixed at 500 for each experiment.

%
%

\paragraph{Monocular Encoder Generalization with Diverse Exposure.}
Fig.~\ref{fig:ablation_encoder_generalization} investigates the benefit of exposing the monocular image encoder ($E_{image}$) to diverse identities exhibiting similar expressions during its training, leveraging techniques inspired by LivePortrait~\cite{guo2024liveportrait}.
This pre-exposure aids in learning a more robust expression representation that generalizes better.
The figure compares results driven by an in-the-wild image (right): left shows animation without this diverse pre-exposure, while center shows our method with it.
This diverse training helps achieve more accurate expression alignment when driving the avatar with in-the-wild images of unseen identities and varied conditions.
\paragraph{Geometric Accuracy.}  
To further validate the robustness of our approach, we report quantitative geometry metrics on held-out speaker data (Tab.~\ref{tab:quantitative_comparison_lve_mve_fdd}) on a subject from the Multiface dataset \cite{wuu2023multifacedatasetneuralface}.  
For fair comparison across different 3D mesh topologies (FLAME, Ava-256), we resample all meshes to the Sapiens~\cite{khirodkar2024_sapiens} landmark topology.  
Our method achieves lower lip vertex error (LVE) and mean vertex error (MVE), as well as a strong FDD score, demonstrating that UHAP not only drives photorealistic appearance but also preserves accurate geometric motion compared to ground-truth facial dynamics.
\paragraph{Additional qualitative results}
We further provide qualitative results on subjects from the HQ3DAvatar~\cite{teotia2023hq3davatar} and RenderMe-360~\cite{pan2024renderme} datasets.  
As shown in Fig.~\ref{fig:extra_results}, our method takes as input expression data from multiple views for each subject, and synthesizes high-fidelity audio-driven facial animation, consistently across all the subjects.  
\begin{table}[t]
\small
  \centering
  \caption{Quantitative comparison with SOTA audio-driven avatar methods on landmark and facial dynamics distance metrics.}
  \label{tab:quantitative_comparison_lve_mve_fdd}
  \sisetup{table-align-text-post=false} 
  \begin{tabular}{@{}lS[table-format=1.2]S[table-format=1.2]S[table-format=1.4]@{}}
    \toprule
    {Method} & {LVE [mm] $\downarrow$} & {MVE [mm] $\downarrow$} & {FDD $\downarrow$} \\
    \midrule
    CodeTalker   & 4.13 & 4.25 & 0.4902 \\
    FaceFormer   & 3.92 & 4.16 & 0.4390 \\
    FaceDiffuser & 3.46 & 3.98 & 0.4060 \\
    \midrule
    \textbf{Ours}    & \textbf{3.01} & \textbf{3.15} & \textbf{0.1848} \\
    \bottomrule
  \end{tabular}
\end{table}

\begin{figure}[h!]
\centering
\includegraphics[width=0.8\linewidth]{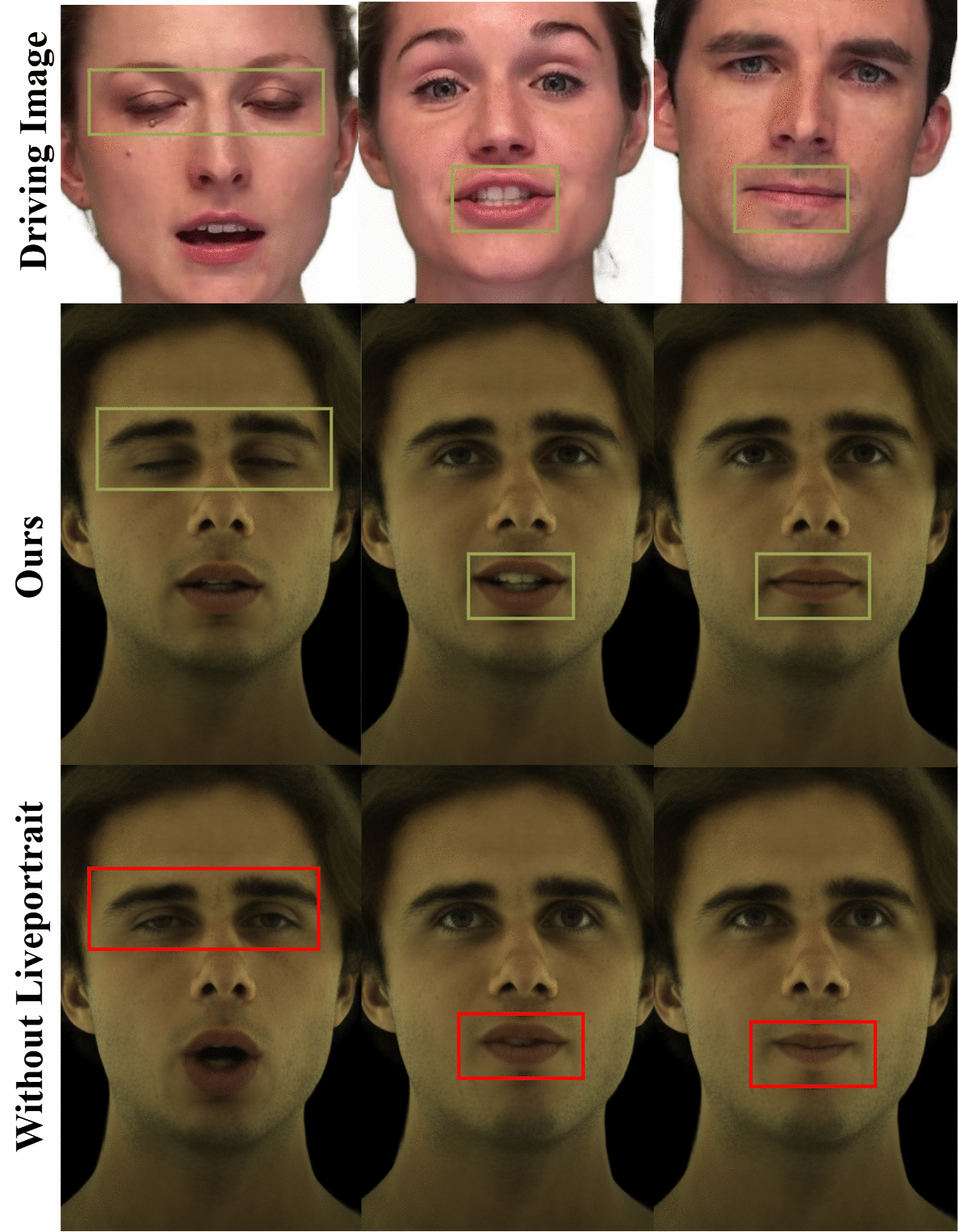} 
\caption{Ablation on monocular encoder generalization using LivePortrait2D~\cite{guo2024liveportrait} inspired diverse pre-exposure: Without diverse pre-exposure (left), Ours (center), Driving Image (right).}
\label{fig:ablation_encoder_generalization} 
\end{figure}

\begin{figure}[h!]
\centering
\includegraphics[width=0.9\linewidth]{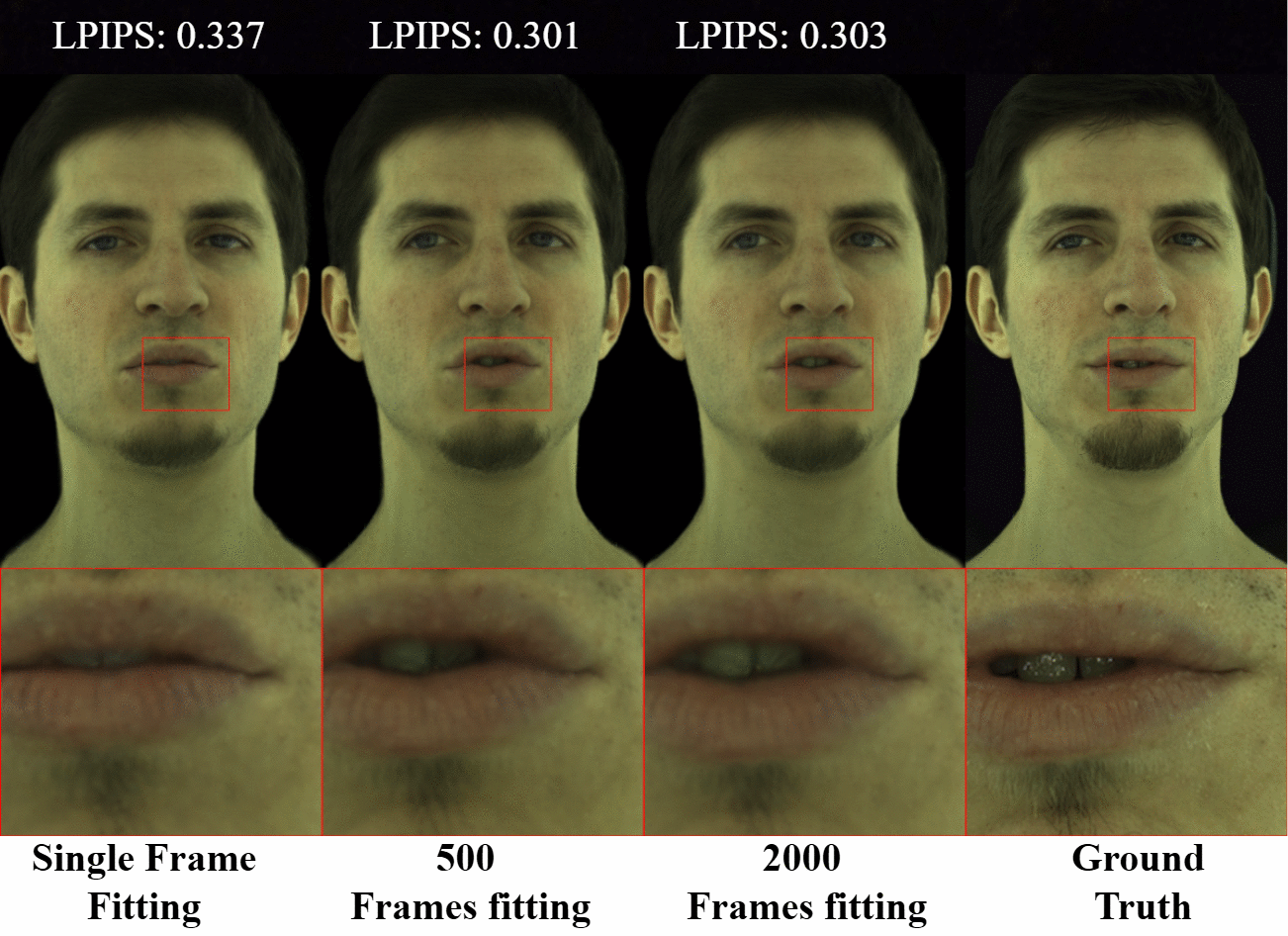} 
\caption{Ablation on number of frames used to fine-tune UHAP.}
\label{fig:frame_ablation} 
\end{figure}

\begin{figure}[h!]
\centering
\includegraphics[width=0.9\linewidth]{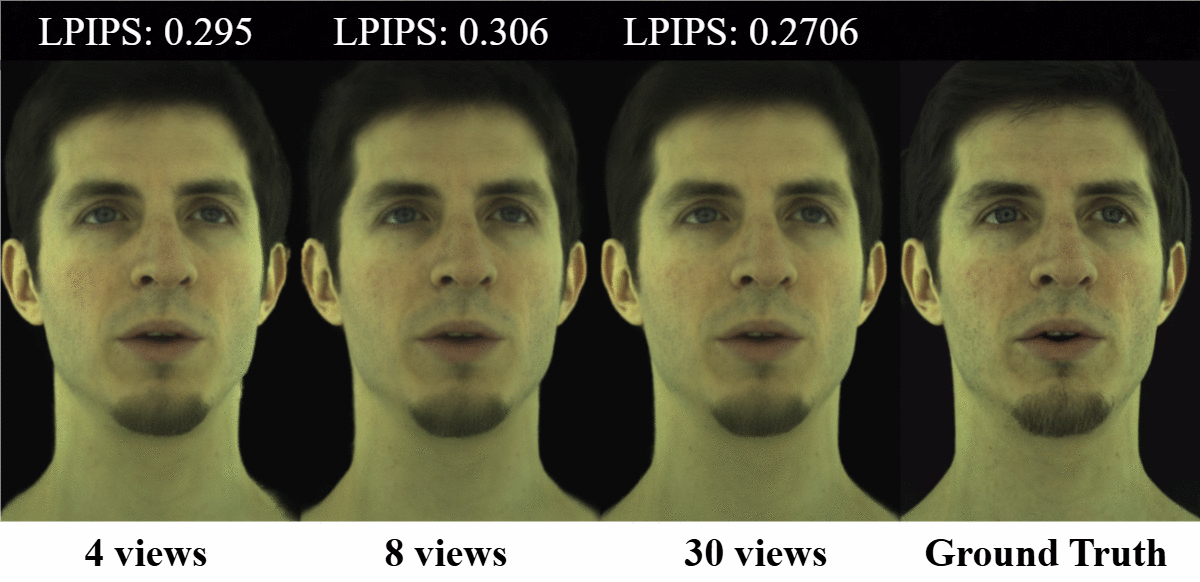} 
\caption{Ablation on number of input views used to fine-tune UHAP.}
\label{fig:view_ablation} 
\end{figure}
\begin{figure*}[htb]
\centering
\includegraphics[width=0.9\textwidth]{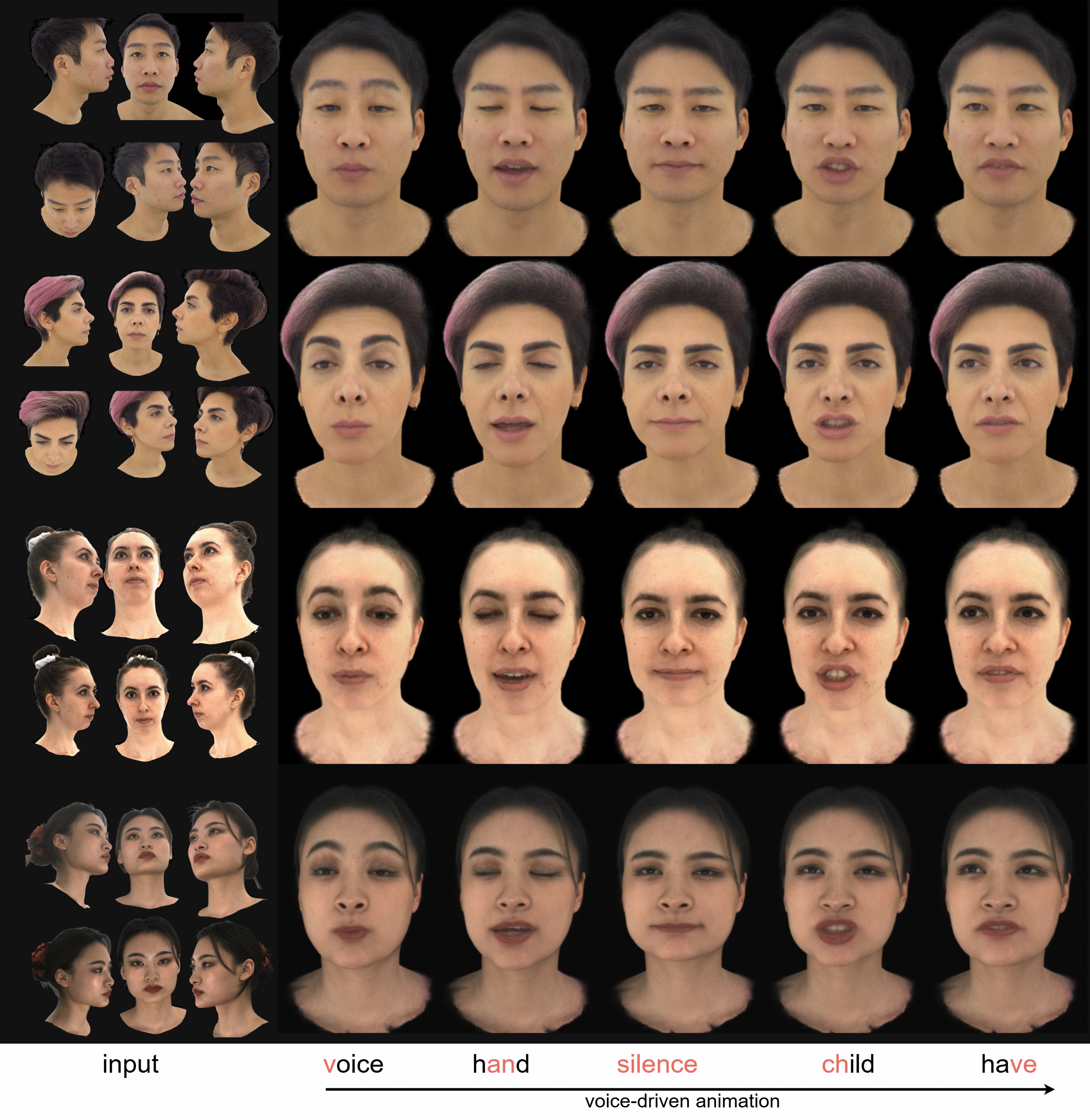} 
\caption{Additional audio-driven qualitative results on sparse input data from subjects of HQ3DAvatar \cite{teotia2023hq3davatar} (top 2 rows) and RenderMe-360 \cite{pan2024renderme} (bottom 2 rows) datasets. Our method produces high-fidelity facial animation across diverse identities.}
\label{fig:extra_results}
\end{figure*}


\section{Implementation Details}
\label{sec:implementation_details}

This section details the architectures of the core components of our framework: the Universal Head Avatar Prior (UHAP), the Monocular Expression Encoder ($E_{image}$), and the Audio-to-Expression Diffusion Model ($\mathcal{G}_{\theta}$), as well as training specifics.

\subsection{UHAP Components}
\label{sec:uhap_implementation_details_subsection}

The UHAP, is composed of several interconnected neural network modules. These modules are responsible for encoding expressions ($E_{exp}$), representing identity ($\mathbf{Z_{id}}$), and decoding these into a full 3D Gaussian avatar via $\mathcal{D}_{UHAP}$ (which includes $\mathcal{D}_{neut}$, $\mathcal{D}_{guide}$, and $\mathcal{D}_{ga}$). Below, we detail their architectures, with layer configurations summarized in the accompanying tables.

\subsubsection{Expression Encoder ($E_{exp}$)}
The Expression Encoder $E_{exp}$ processes UV-parameterized texture difference maps ($\Delta T_{exp}$) and geometry difference maps ($\Delta G_{exp}$) to produce the parameters of the expression code $\mathbf{Z_{exp}}$. The input UV maps are of size $512 \times 512$ with 3 channels. As detailed in Table~\ref{tab:cnn_encoder_posmap} (module `CNNEncoderPosmap`), the encoder consists of a series of 8 convolutional blocks. Each block applies a 2D convolution, followed by a LeakyReLU activation and downsampling, progressively reducing the spatial resolution from $512 \times 512$ down to $2 \times 2$ while adjusting channel depth. The final $256 \times 2 \times 2$ feature map is flattened and passed through a fully connected layer to output the 256-dimensional parameters ($\boldsymbol{\mu}_{exp}, \boldsymbol{\sigma}_{exp}$) for $\mathbf{Z_{exp}}$.

\begin{table}[h!]
\centering
\caption{Architecture of the UHAP Expression Encoder $E_{exp}$. C = Conv2dUB; LR = LeakyReLU; DS = down‐sample; FC = Fully Connected.}
\label{tab:cnn_encoder_posmap}
\resizebox{\linewidth}{!}{%
\begin{tabular}{|l|c|c|}
\hline
            & \textbf{Channels / Resolution} & \textbf{Operation}      \\
\hline
Input       & 3 @ $512\times512$             & N/A                     \\
\hline
Block 1     & (3$\to$32) @ $512\to256$        & (C, LR, DS)             \\
Block 2     & (32$\to$32) @ $256\to128$       & (C, LR, DS)             \\
Block 3     & (32$\to$64) @ $128\to64$        & (C, LR, DS)             \\
Block 4     & (64$\to$64) @ $64\to32$         & (C, LR, DS)             \\
Block 5     & (64$\to$128) @ $32\to16$        & (C, LR, DS)             \\
Block 6     & (128$\to$128) @ $16\to8$        & (C, LR, DS)             \\
Block 7     & (128$\to$256) @ $8\to4$         & (C, LR, DS)             \\
Block 8     & (256$\to$256) @ $4\to2$         & (C, LR, DS)             \\
\hline
Output      & $256 \text{ @ } 2\times2 \to \text{FC}(256)$     & Flatten + FC            \\
\hline
\end{tabular}%
}
\end{table}

\subsubsection{Identity Representation ($\mathbf{Z_{id}}$)}
The identity code $\mathbf{Z_{id}}$ is a learnable embedding vector for each subject. For $N_{ids}$ unique identities in the training set, an embedding table of size $N_{ids} \times D_{id}$ is maintained, where $D_{id}=512$ is the dimension of the identity latent code. The corresponding 512-dimensional vector $\mathbf{Z_{id}}$ is retrieved via lookup (module `IdentityLatentCode`, Table~\ref{tab:identity_latent_code}).

\begin{table}[h!]
\centering
\caption{Identity Latent Code Module.}
\label{tab:identity_latent_code}
\resizebox{0.7\linewidth}{!}{%
\begin{tabular}{|l|c|c|}
\hline
            & \textbf{Embedding Size}         & \textbf{Operation}      \\
\hline
Input       & $N_{ids}$ indices               & N/A                     \\
\hline
Lookup      & ($N_{ids} \times D_{id}$)       & Embedding lookup        \\
\hline
Output      & $D_{id}$                     & per‐subject vector      \\
\hline
\end{tabular}%
}
\end{table}

\subsubsection{Neutral Decoder ($\mathcal{D}_{neut}$)}
The Neutral Decoder $\mathcal{D}_{neut}$ takes the $D_{id}$-dimensional identity code $\mathbf{Z_{id}}$ as input and generates the identity-specific feature maps $\mathbf{f}_{neut}$. These maps comprise two sets of multi-scale bias maps: $\mathbf{f}_{neut,geo}$ for geometry and $\mathbf{f}_{neut,app}$ for appearance. Each set is produced by dedicated generators. As detailed in Table~\ref{tab:deconv_bias_map_generator}, each generator processes the 512-dim $\mathbf{Z_{id}}$ through a series of deconvolutional blocks. This produces a pyramid of 9 bias maps, where each map in the pyramid has a progressively larger spatial resolution (from $4\times4$ up to $1024\times1024$) and a corresponding number of channels as specified in the table. These $\mathbf{f}_{neut}$ maps are then injected (by element-wise add) at matching scales into the Gaussian Avatar Decoder $\mathcal{D}_{ga}$ to provide identity-specific conditioning.

\begin{table}[h!]
\centering
\caption{Architecture of Bias Map Generators for $\mathbf{f}_{neut}$. The "Channels" for "Blocks (Deconv)" lists the output channels for each of the 9 generated bias maps corresponding to the increasing resolutions.}
\label{tab:deconv_bias_map_generator}
\resizebox{\linewidth}{!}{%
\begin{tabular}{|l|c|c|}
\hline
            & \textbf{Channels}              & \textbf{Resolutions}      \\
\hline
Input       & $D_{id}$ (512)                            & N/A                       \\
\hline
Blocks (Deconv) & [256,256,128,128,64,64,32,16,3] & $4\times4 \to 8\times8 \to \dots \to 1024\times1024$  \\
\hline
Output      & 9 bias maps                    & multi‐scale               \\
\hline
\end{tabular}%
}
\end{table}

\subsubsection{Guide Mesh Decoder ($\mathcal{D}_{guide}$)}
The Guide Mesh Decoder $\mathcal{D}_{guide}$ predicts per-vertex displacements for a canonical template mesh of $N_{verts}=7306$ vertices. It processes a 768-dimensional vector, formed by concatenating $\mathbf{Z_{id}}$ ($D_{id}=512$) and $\mathbf{Z_{exp}}$ ($D_{exp}=256$), through an MLP with LeakyReLU activations, as detailed in Table~\ref{tab:mlp_decoder}. The output is reshaped to $N_{verts} \times 3$ displacement vectors, which are added to the canonical mesh vertices to yield $\mathbf{\hat{v}_p}$.

\begin{table}[h!]
\centering
\caption{Architecture of the Guide Mesh Decoder $\mathcal{D}_{guide}$. FC = Fully Connected; LR = LeakyReLU.}
\label{tab:mlp_decoder}
\resizebox{\linewidth}{!}{%
\begin{tabular}{|l|c|c|}
\hline
            & \textbf{Units}                  & \textbf{Operation}      \\
\hline
Input       & $D_{exp}$ (256) + $D_{id}$ (512) = 768     & N/A                     \\
\hline
Block 1     & $768 \to 1024$                      & (FC, LR)                \\
Block 2     & $1024 \to 2048$                     & (FC, LR)                \\
Block 3     & $2048 \to 3 \times N_{verts}$  & FC                      \\
\hline
Output      & $N_{verts}\times3$     & reshape to offsets      \\
\hline
\end{tabular}%
}
\end{table}

\subsubsection{Gaussian Avatar Decoder ($\mathcal{D}_{ga}$)}
The Gaussian Avatar Decoder $\mathcal{D}_{ga}$ synthesizes the final 3D Gaussian parameters through two sub-decoders: a view-independent decoder $\mathcal{D}_{vi}$ for geometry-related attributes and an appearance decoder $\mathcal{D}_{rgb}$ for view-dependent color. Both are conditioned on $\mathbf{Z_{id}}$, $\mathbf{Z_{exp}}$, and $\mathbf{f}_{neut}$.

\noindent\textbf{View-Independent Decoder ($\mathcal{D}_{vi}$):}
This component (architecture in Table~\ref{tab:decoder_vi}) predicts view-independent Gaussian parameters: positional offsets $\delta \mathbf{t}_k$, rotations $\mathbf{q}_k$, scales $\mathbf{s}_k$, and opacity $o_k$. Input is the concatenated $\mathbf{Z_{id}}$ and $\mathbf{Z_{exp}}$. An initial FC layer projects this to a $256 \times 8 \times 8$ feature map. Subsequent blocks perform bias injection (using $\mathbf{f}_{neut, geo}$), transposed convolution for upsampling, and LeakyReLU activation, producing a map (e.g., $11 \times 512 \times 512$) which is reshaped to provide parameters for each of the $N_g$ Gaussians.

\begin{table}[h!]
\centering
\caption{Architecture of the View-Independent Decoder ($\mathcal{D}_{vi}$). Bias-inject uses $\mathbf{f}_{neut,geo}$; WN = weight‐norm; C = Transposed Conv; LR = LeakyReLU.}
\label{tab:decoder_vi}
\resizebox{\linewidth}{!}{%
\begin{tabular}{|l|c|c|}
\hline
            & \textbf{Channels / Feature Map Size}        & \textbf{Operation}      \\
\hline
Input       & 768 ($D_{exp}$ + $D_{id}$)                & N/A                     \\
\hline
Block 1     & FC $\to$ 256 @ $8\times8$                   & (FC+WN, LR)             \\
Block 2     & (256$\to$128) @ $8\times8 \to 16\times16$               & (Bias‐inject, C, LR)    \\
Block 3     & (128$\to$128) @ $16\times16 \to 32\times32$              & (Bias‐inject, C, LR)    \\
Block 4     & (128$\to$64)  @ $32\times32 \to 64\times64$              & (Bias‐inject, C, LR)    \\
Block 5     & (64$\to$64)   @ $64\times64 \to 128\times128$             & (Bias‐inject, C, LR)    \\
Block 6     & (64$\to$32)   @ $128\times128 \to 256\times256$            & (Bias‐inject, C, LR)    \\
Block 7     & (32$\to$16)   @ $256\times256 \to 512\times512$            & (Bias‐inject, C, LR)    \\
Block 8     & (16$\to$11)   @ $512\times512$            & (Bias‐inject, C)        \\
\hline
Output      & $11\times512\times512 \to N_g$ Gaussian params     & split and reshape       \\
\hline
\end{tabular}%
}
\end{table}

\noindent\textbf{Appearance Decoder ($\mathcal{D}_{rgb}$):}
This component (architecture in Table~\ref{tab:decoder_rgb}) predicts view-dependent 3-channel RGB color $\mathbf{c}_k \in \mathbb{R}^3$. It takes concatenated $\mathbf{Z_{id}}$, $\mathbf{Z_{exp}}$, and viewpoint features $V$ as input. Similar to $\mathcal{D}_{vi}$, it uses an FC layer and upsampling blocks with bias injection (using $\mathbf{f}_{neut, app}$), culminating in a $3 \times 512 \times 512$ map for the RGB color $\mathbf{c}_k$ for each Gaussian.

\begin{table}[h!]
\centering
\caption{Architecture of the Appearance Decoder ($\mathcal{D}_{rgb}$). Bias‐inject uses $\mathbf{f}_{neut,app}$; WN = weight‐norm; C = Transposed Conv; LR = LeakyReLU.}
\label{tab:decoder_rgb}
\resizebox{\linewidth}{!}{%
\begin{tabular}{|l|c|c|}
\hline
            & \textbf{Channels / Feature Map Size}               & \textbf{Operation}               \\
\hline
Input       & $\sim$771 ($D_{exp}$ + $D_{id}$ + $D_{view}$) & N/A                            \\
\hline
Block 1     & FC $\to$ 256 @ $8\times8$                           & (FC+WN, LR)                     \\
Block 2     & (256$\to$128) @ $8\times8 \to 16\times16$                         & (Bias‐inject, C, LR)            \\
Block 3     & (128$\to$128) @ $16\times16 \to 32\times32$                        & (Bias‐inject, C, LR)            \\
Block 4     & (128$\to$64) @ $32\times32 \to 64\times64$                         & (Bias‐inject, C, LR)            \\
Block 5     & (64$\to$64) @ $64\times64 \to 128\times128$                         & (Bias‐inject, C, LR)            \\
Block 6     & (64$\to$32) @ $128\times128 \to 256\times256$                        & (Bias‐inject, C, LR)            \\
Block 7     & (32$\to$16) @ $256\times256 \to 512\times512$                        & (Bias‐inject, C, LR)            \\
Block 8     & (16$\to 3$) @ $512\times512$                        & (Bias‐inject, C)                \\
\hline
Output      & $3 \times 512 \times 512 \to N_g$ RGB color                           & reshape       \\
\hline
\end{tabular}%
}
\end{table}

The outputs from these decoders constitute the full set of parameters for the $N_g$ 3D Gaussians, used for rendering the final avatar image.

\subsection{Monocular Expression Encoder ($E_{image}$)}
\label{sec:eimage_implementation_details}
The Monocular Expression Encoder $E_{image}$, responsible for predicting the 256-dimensional expression code $\mathbf{\hat{Z}_{exp}}$ from a single input image. It employs a two-branch architecture inspired by Live3DPortrait \cite{trevithick2023}. However, instead of predicting triplanes, our $E_{image}$ directly regresses the latent expression code $\mathbf{\hat{Z}_{exp}}$ compatible with our UHAP.

The encoder processes the input image $I_i$ (e.g., $512 \times 512 \times 3$) through two parallel pathways:
\begin{itemize}
    \item \textbf{Low-Resolution Branch:} A truncated ResNet34~\cite{DBLP:journals/corr/HeZRS15} acts as a feature extractor (\texttt{low\_feat\_extractor}), producing features ($512 \times H/32 \times W/32$). These are passed through a $1 \times 1$ convolution (\texttt{low\_conv}) to reduce channel dimensionality (e.g., to 128). An \texttt{OverlapPatchEmbed} module then converts these features into patch embeddings of dimension 256. These patch embeddings are processed by a \texttt{TransformerBlock} (\texttt{vit\_block}) and reshaped back into a 2D feature map ($256 \times H/32 \times W/32$).
    \item \textbf{High-Resolution Branch:} A simple CNN, \texttt{HighResEncoder} (two Conv2D-ReLU layers reducing $512 \times 512 \times 3 \to 128 \times 128 \times 64$), extracts high-frequency details. These features are also passed through a $1 \times 1$ convolution to match channel dimensions (to 128) and then bilinearly interpolated to the same spatial dimensions ($H/32 \times W/32$) as the output of the low-resolution branch.
\end{itemize}
The feature maps from both branches are concatenated channel-wise (e.g., \(256+128=384\) channels) and fused using a \(3 \times 3\) convolution, reducing channels back to 256. This fused feature map is then flattened and processed by a ViT-decoder \cite{dosovitskiy2021imageworth16x16words}. The output of this decoder is followed by adaptive average pooling and a linear projection layer to output the final 256-dimensional expression code $\mathbf{\hat{Z}_{exp}}$.

\subsection{Audio-to-Expression Diffusion Model ($\mathcal{G}_{\theta}$)}
\label{sec:audio_model_implementation_details}
The audio-to-expression diffusion model $\mathcal{G}_{\theta}$, is based on the Transformer architecture from \cite{ng2024audiophotorealembodimentsynthesizing}. It consists of $L=6$ layers, with each multi-head attention mechanism employing $H=8$ heads. During training, for classifier-free guidance (to avoid overfittin), conditioning signals (audio features and predicted lip vertices) are masked with a probability of $p_{cond\_mask}=0.25$.

\noindent\textbf{Inference.} At inference time, the expression code sequence $\mathbf{\hat{Z}^{0}_{exp}}$ is generated by iteratively denoising a random Gaussian noise sample for $N_{diff}=500$ diffusion steps. To handle long audio sequences and maintain temporal coherence, we adopt a windowed generation approach. The audio is processed in overlapping windows (120 frames). For each window beyond the first, the last 30 frames of the previously generated expression sequence are normalized using pre-computed dataset statistics (mean and standard deviation of expression codes) and provided context to condition the generation of the current window. Only the new, non-overlapping portion of each generated window is retained, ensuring smooth transitions across the full sequence.

\subsection{Hyperparameters and Training Resources}
\label{sec:hyperparameters_training}
The training of our UHAP model and the personalization fine-tuning stage involve several loss terms weighted by hyperparameters. These weights balance the contribution of each term to the overall objective. Table~\ref{tab:uhap_loss_weights} details the hyperparameter weights $\lambda_{(\cdot)}$ for the UHAP training objective $\mathcal{L}_{UHAP}$. Table~\ref{tab:fit_loss_weights} specifies the weights $\alpha_i$ for the personalization fitting objective $\mathcal{L}_{fit}$. The values presented are indicative and typically determined through empirical validation.

\begin{table}[h!]
\centering
\caption{Hyperparameter weights for the UHAP training objective $\mathcal{L}_{UHAP}$.}
\label{tab:uhap_loss_weights}
\begin{tabular}{lc}
\toprule
\textbf{Hyperparameter} & \textbf{Value} \\
\midrule
$\lambda_{rec}$ & $1.0$ \\
$\lambda_{neut}$ & $1e\text{-}3$ \\
$\lambda_{KL}$ & $1e\text{-}2$ \\
$\lambda_{geo}$ & $1e\text{-}3$ \\
$\lambda_{perc}$ & $1e\text{-}2$ \\
$\lambda_{reg\_id}$ & $1e\text{-}3$ \\
$\lambda_{reg\_gauss}$ & $1e\text{-}3$ \\
\bottomrule
\end{tabular}
\end{table}

\begin{table}[h!]
\centering
\caption{Hyperparameter weights for the personalization objective $\mathcal{L}_{fit}$.}
\label{tab:fit_loss_weights}
\begin{tabular}{lc}
\toprule
\textbf{Hyperparameter} & \textbf{Value} \\
\midrule
$\alpha_1$ & $1$ \\
$\alpha_2$  & $1e\text{-}2$ \\
$\alpha_3$& $1e\text{-}3$ \\
$\alpha_4$ & $1e\text{-}2$ \\

\bottomrule
\end{tabular}
\end{table}

\noindent\textbf{Training Time and Resources.}
The Universal Head Avatar Prior (UHAP) is trained for a total of 300k iterations on 4 NVIDIA A40 GPUs (with a batch size of 1 per GPU). The Monocular Expression Encoder ($E_{image}$) is subsequently trained for 100k iterations. The audio-to-expression diffusion model ($\mathcal{G}_{\theta}$) is trained for 200k iterations on a single NVIDIA A40 GPU.



\end{document}